\documentclass[10pt,twocolumn,letterpaper]{article}

\usepackage[pagenumbers]{cvpr} % To force page numbers, e.g. for an arXiv version

\newcommand{\TODO}[1]{\textbf{\color{red}[TODO: #1]}}
\graphicspath{{figs/}}
\renewcommand{\TODO}[1]{}

\definecolor{cvprblue}{rgb}{0.21,0.49,0.74}
\usepackage{xspace}
\usepackage[pagebackref,breaklinks,colorlinks,allcolors=cvprblue]{hyperref}
\newcommand{\modelname}{\textsc{NoRD}\xspace}
\usepackage{xcolor}
\usepackage{ifthen}
\usepackage{makecell}

\usepackage{pifont}

\newcommand{\good}{\textcolor{green!60!black}{\ding{51}}} % tick
\newcommand{\bad}{\textcolor{red!70!black}{\ding{55}}}     % cross

\newboolean{shownotes}
\setboolean{shownotes}{true} % set to false to hide all notes

\newcommand{\note}[3]{%
  \ifthenelse{\boolean{shownotes}}{%
    {\color{#1}\textbf{[#2: #3]}}%
  }{}%
}

\usepackage{mathtools}
\def\paperID{*****} % *** Enter the Paper ID here
\def\confName{CVPR}
\def\confYear{2026}

\title{\modelname: A Data-Efficient Vision-Language-Action Model that Drives without Reasoning}

\author{
Ishaan Rawal\textsuperscript{1,2}\thanks{Work done during an internship at Applied Intuition.} \quad
Shubh Gupta\textsuperscript{1} \quad
Yihan Hu\textsuperscript{1} \quad
Wei Zhan\textsuperscript{1,3}\thanks{Corresponding author. Email: \texttt{wei.zhan@applied.co}} \\[2mm]
\textsuperscript{1}Applied Intuition \quad
\textsuperscript{2}Texas A\&M University \quad
\textsuperscript{3}UC Berkeley
}

\begin{document}

\maketitle
\begin{abstract}
Vision-Language-Action (VLA) models are advancing autonomous driving by replacing modular pipelines with unified end-to-end architectures. However, current VLAs face two expensive requirements: (1) massive dataset collection, and (2) dense reasoning annotations. In this work, we address both challenges with \modelname (\textbf{No} \textbf{R}easoning for \textbf{D}riving). Compared to existing VLAs, \modelname achieves competitive performance while being fine-tuned on $<$60\% of the data and no reasoning annotations,  resulting in 3$\times$ fewer tokens. We identify that standard Group Relative Policy Optimization (GRPO) fails to yield significant improvements when applied to policies trained on such small, reasoning-free datasets. We show that this limitation stems from difficulty bias, which disproportionately penalizes reward signals from scenarios that produce high-variance rollouts within GRPO. \modelname overcomes this by incorporating Dr.~GRPO, a recent algorithm designed to mitigate difficulty bias in LLMs. As a result, \modelname achieves competitive performance on Waymo and NAVSIM with a fraction of the training data and no reasoning overhead, enabling more efficient autonomous systems.

% trained on $<$50\% of the data and without reasoning annotations, resulting in 15× fewer tokens while maintaining high accuracy. \modelname achieves this efficiency by shifting more of the learning from finetuning to the Reinforcement Learning (RL) post-training stage. Existing VLAs utilize a GRPO-based post-training stage that contributes minimally to an already well-performing finetuned model trained on large datasets with reasoning. We demonstrate that GRPO is ineffective at improving weaker models trained on less data without reasoning.  We identify that this is due to the difficulty bias in GRPO, where reward signal from very easy or very hard samples are under-weighted and learning is dominated by samples with low reward variance. \modelname addresses this challenge by replacing the GRPO in the RL-stage with DrGRPO, a recent algorithm designed to mitigate difficulty bias in LLMs. We find that DrGRPO-based post-training significantly improves weaker the model with fewer samples. As a result, \modelname achieves competitive performance on Waymo and NAVSIM without reasoning or additional inputs, enabling scalable, data-efficient training and fast inference.
\end{abstract}          
\section{Introduction}
\label{sec:intro}
\begin{figure}[t]
  \centering
   \includegraphics[width=1.0\linewidth]{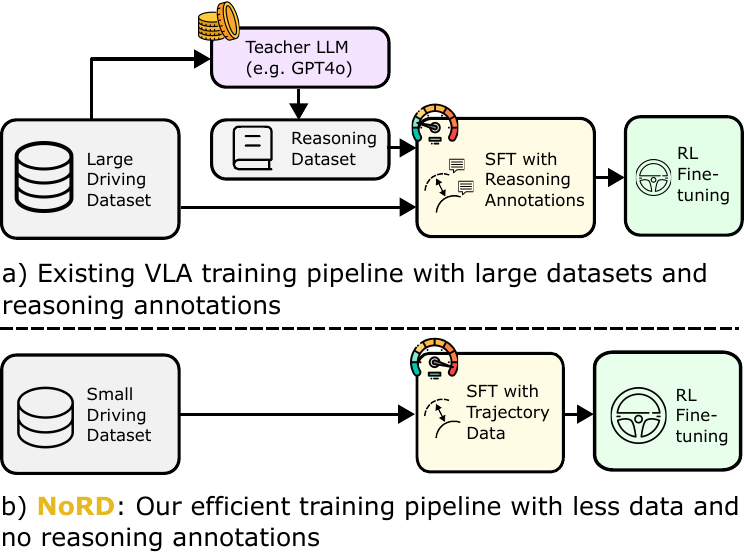}

   \caption{\textbf{Comparison of VLA training pipelines.} (a) Existing approaches depend on large-scale reasoning data generation, followed by extensive SFT and RL fine-tuning. (b) In contrast, \modelname directly utilizes a small-scale driving dataset for SFT, and performs RL fine-tuning tailored for weak SFT policy, enabling data-efficient learning without reasoning supervision.}
   \label{fig:nord_intro}
\end{figure}
% \shubh{Topic: VLAs are important for AD}
The prevailing paradigm for end-to-end autonomous driving is increasingly shifting toward Vision-Language-Action (VLA) models. The dominant training methodology for these models is a two-stage training pipeline: (1) Supervised Fine-Tuning (SFT) on large-scale datasets with detailed, natural language Chain-of-Thought (CoT) reasoning annotations~\cite{zhou2025autovla, wang2025alpamayo, yuan2025autodrive}, followed by (2) a Reinforcement Learning (RL) stage to align outputs with driving metrics, for which Group Relative Policy Optimization (GRPO)~\cite{guo2025deepseek} has been widely adopted~\cite{zhou2025autovla,rowe2025poutine}.

% \shubh{Topic: limitations of cot then rl}
While this paradigm has achieved state-of-the-art performance on complex driving benchmarks~\cite{rowe2025poutine}, its reliance on both massive data and dense reasoning introduces three non-scalable costs:
\begin{enumerate}
    \item \textit{Data cost} of collecting and curating vast quantities of specialized driving scenarios
    \item \textit{Annotation cost} from generating high-quality reasoning traces for this data 
    \item \textit{Training and inference cost} from resulting reasoning tokens, increasing training time and creating inference latency that is impractical for real-world deployment
\end{enumerate}
This motivates a natural hypothesis: \textbf{Can we achieve competitive performance on driving benchmarks while being both reasoning-free and data-efficient?} 

% \shubh{Topic: our reasoning-free hypothesis}
This investigation is supported by two distinct lines of work. First, recent studies provide a theoretical motivation by questioning the necessity of explicit reasoning, suggesting it may be a byproduct of planning rather than a causal determinant~\cite{song2025more}. Second, existing work on end-to-end models like EMMA~\cite{hwang2025emma} and SimLingo~\cite{renz2025simlingo}, provides an empirical precedent by achieving strong performance on nominal benchmarks without reasoning. We therefore investigate if this data-efficient, reasoning-free approach can be extended to the more challenging benchmarks (e.g. NAVSIM~\cite{Dauner2024NEURIPS}, WaymoE2E~\cite{waymo2025e2e}) that are currently dominated by their reasoning-centric counterparts.

% \shubh{Topic: Challenge of reasoning-free}
We initially trained a reasoning-free \modelname{\textsc{-base}} VLA (based on Qwen-2.5VL-3B-Instruct~\cite{bai2025qwen2}) using only SFT on 80,000 NAVSIM training samples; a greater than 60\% reduction in data compared to state-of-the-art reasoning-based models~\cite{zhou2025autovla}. This model was then post-trained with GRPO to optimize the PDM score~\cite{Dauner2024NEURIPS}. 

However, this data-efficient, reasoning-free model achieves scores significantly lower than reasoning-based baselines ($>$12-point difference), and post-training with GRPO only results in a meager improvement (+0.67\%). This initial failure creates the illusion that reasoning data is a necessary component for achieving high performance.

% \shubh{Topic: GRPO is the problem}

In this work, we argue that this conclusion is premature. We posit that the failure lies not in the reasoning-free SFT policy, but rather in the interaction between the policy optimization method (GRPO) and the reward landscape. We find that the complex, sparse reward signals from driving benchmarks (like PDM score from NAVSIM or the RFS from WaymoE2E) induce a highly polarized distribution of intra-group rewards. A significant mass of these mean rewards is clustered at the extremes (i.e., near 0 or 1), and correspond to rollouts with low variance. Conversely, the remaining scenarios, which yield intermediate mean rewards, are characterized by high-variance rollouts. When GRPO is applied in this landscape to a weaker, data-efficient SFT policy like \modelname{\textsc{-base}}, the resulting learning signal disproportionately penalizes the intermediate-mean (high-variance) scenarios, impeding effective optimization.

We are the first to identify that the failure to optimize weak SFT mode is caused by polarized intra-group reward landscape, and that it stems from \textit{difficulty bias}, which has also been observed in LLM reasoning domain~\cite{liu2025understanding, li2025disco}. Based on our analysis, we propose to mitigate this bias by using Dr.~GRPO~\cite{liu2025understanding}, an existing policy optimization algorithm specifically designed to address this flaw. We demonstrate that by applying Dr.~GRPO as a drop-in replacement, our reasoning free VLA, \modelname, can be successfully trained.

Our key contributions are as follows:
\begin{enumerate}
    \item We are the first to identify that the failure of reasoning-free and data-efficient VLA training for autonomous driving is an instance of difficulty bias, triggered by the combination of a weak SFT policy and complex driving metrics.
    \item We empirically characterize this failure, showing that the data-efficient SFT policy induces a polarized reward distribution that deprives GRPO of a learning signal.
    \item We propose using Dr.~GRPO as a drop-in replacement to train \modelname, a data-efficient, reasoning-free VLA, and are the first to validate this policy optimization method in the autonomous driving domain (see \cref{fig:nord_intro}).
    \item We demonstrate performance competitive with the state-of-the-art on the NAVSIM and WaymoE2E benchmarks without using any reasoning annotations and at least 60\% less data than reasoning VLAs, while improving on inference time, proving the viability of our approach.
\end{enumerate}

% VLAs represent the forefront of end-to-end autonomous driving. They leverage large-scale pre-training for robust decision making. SOTA methods typically employ a 2-stage training paradigm, SFT with expensive COT reasoning annotations followed by RL via GRPO to align with driving metrics. This reliance on COT annotations, however, is a major bottleneck, demanding expensive data generation from a larger "teacher" model and incurring significant inference latency. This renders these approaches impractical for deployment on real-world vehicles.

% We investigate a "reasoning-free" paradigm, optimizing Qwen-2.5VL-3B-Instruct using SFT on trajectories followed directly by DrGRPO optimization on complex driving metrics (PDM score, RFS). 

\section{Related Works}
\label{sec:related_works}

\noindent\textbf{Reasoning-based VLAs.} Several works have incorporated high-level reasoning into the control loop, including hybrid architectures like ORION~\cite{fu2025orion}, unified transformers like AutoVLA~\cite{zhou2025autovla}, and a wide variety of reasoning strategies, such as retrieval-augmented CoT~\cite{drivingvqa2025}, spatio-temporal reasoning~\cite{zeng2025futuresightdrive}, multi-agent reasoning~\cite{qian2025agentthnk, zheng2025driveagentr1}, and models combining memory and tool use~\cite{li2025recogdrive, liu2025reasonplan}. While this approach has achieved state-of-the-art performance on complex driving benchmarks~\cite{rowe2025poutine}, its reliance on large-scale, specialized reasoning datasets~\cite{wang2025alpamayo, chi2025impromptu, yuan2025autodrive} and the high inference latency of CoT generation~\cite{luo2025adathinkdrive, liao2025cotdrive} motivate the exploration of alternatives.
\vspace{0.5em}
\newline
\noindent\textbf{Reasoning-Planning Disconnect.}\ The high cost of reasoning-centric models has spurred an inquiry into their necessity, with recent work questioning whether the model's reasoning improves its planning output. One study~\cite{song2025more}, proposing a ``Reasoning-Planning Decoupling Hypothesis" demonstrated that textual priors alone can match the performance of full multimodal reasoning models. This skepticism extends to RL post-training, as other works argue that RL does not instill new reasoning capacity but instead optimizes within the SFT model's existing latent distribution~\cite{yue2025does}. These findings motivate our reasoning-free approach and frame our central question: does the failure to align weak SFT models stem from an inherent limitation of these models, or from an optimization failure?
\vspace{0.5em}
\newline
\noindent\textbf{Reasoning-Free VLAs.} A separate line of VLA models operate without explicit reasoning traces. This includes models that map raw sensor data directly to trajectories like EMMA~\cite{hwang2025emma}, SimLingo~\cite{renz2025simlingo}, and S4-Driver~\cite{xie2025s4driver}, as well as generative approaches like ADriver-I~\cite{jia2023adriver1}, DrivingGPT~\cite{chen2025drivinggpt}, and DiffVLA~\cite{jiang2025diffvla}. While these methods have demonstrated strong performance on nominal driving benchmarks like nuScenes~\cite{caesar2020nuscenes}, they have not yet proven competitive on the complex, long-tail benchmarks where reasoning-centric models currently excel.
\vspace{0.5em}
\newline
\noindent\textbf{Data Efficient VLAs.} Our work aims to close the performance gap between reasoning-free and reasoning-based methods while maintaining data efficiency. Many data-efficient VLAs in broader domains~\cite{yang2025egovla, fan2025diffusion, deng2025graspvla, wen2025diffusionvla} mitigate data scarcity by leveraging massive external out-of-domain datasets. We instead focus on the distinct and more challenging problem of training a competitive model using only small-sized specialized, in-domain driving data.
\vspace{0.5em}
\newline
\noindent\textbf{Mitigating Difficulty Bias.} The literature to mitigate the difficulty bias in GRPO, largely from the LLM reasoning domain, is divided into two main strategies. Data-level interventions attempt to manage data before the optimization, using methods like online filtering of saturated or degenerate samples~\cite{cui2025prime, liu2025dapo}, curriculum learning~\cite{parashar2025curriculum}, or advanced sampling~\cite{an2025polaris}. These approaches are designed for binary rewards and are generally computationally infeasible for expensive driving simulations, as they often require multiple rollouts to estimate sample difficulty. In contrast, algorithmic-level interventions modify the optimization algorithm itself. This includes reweighting schemes~\cite{zhou2025daro, zhang2025grpolead}, alternative objectives~\cite{li2025disco, chu2025gpg}, or difficulty-based priors~\cite{chen2025unlockingpotentialdifficultyprior}. Our work incorporates this second strategy. We select Dr.~GRPO~\cite{liu2025understanding}, a lightweight method that directly corrects the bias by identifying and adjusting the specific normalization term in the advantage estimation responsible for it. Dr.~GRPO is thus a prime candidate for our setting, as it avoids the infeasible overhead of data-level methods.

\section{Limitations of GRPO for Data-Efficient Training}
\label{sec:difficulty_bias}

% \begin{figure}[t]
%     \centering
%     \includegraphics[width=0.9\linewidth]{author-kit-CVPR2026-v1-latex-//figures/grpo_steps.png}
%     \caption{\textbf{Evolution of group-mean PDM score during RL fine-tuning with GRPO.} GRPO struggles to optimize samples with high group variance during training, particularly in the range $[0.2$–$0.65]$.}
%     \label{fig:grpo_steps}
% \end{figure}
% \begin{figure}[t]
%         \centering
%         \includegraphics[width=0.9\linewidth]{author-kit-CVPR2026-v1-latex-//figures/drgrpo_steps.png}
%         \caption{\textbf{Evolution of group-mean PDM score during RL fine-tuning with Dr.~GRPO.} Dr.~GRPO effectively optimizes high-variance samples during training, resulting in significant overall performance gains.}
%         \label{fig:drgrpo_steps}
% \end{figure}

% Existing Vision-Language-Action (VLA) models are typically first trained via supervised fine-tuning (SFT), followed by a reinforcement learning (RL) phase. In particular, RL with Verifiable Rewards (RLVR) enables training the fine-tuned model using reward signals that are not end-to-end differentiable, allowing it to learn high-level driving concepts such as compliance with drivable areas, collision avoidance, and preferred trajectory behavior. We hypothesize that shifting more of the learning burden from SFT to RLVR, it should be possible to VLAs efficiently, without needing significant training data or reasoning annotations.
\begin{figure}[t]
  \centering
   \includegraphics[width=0.9\linewidth]{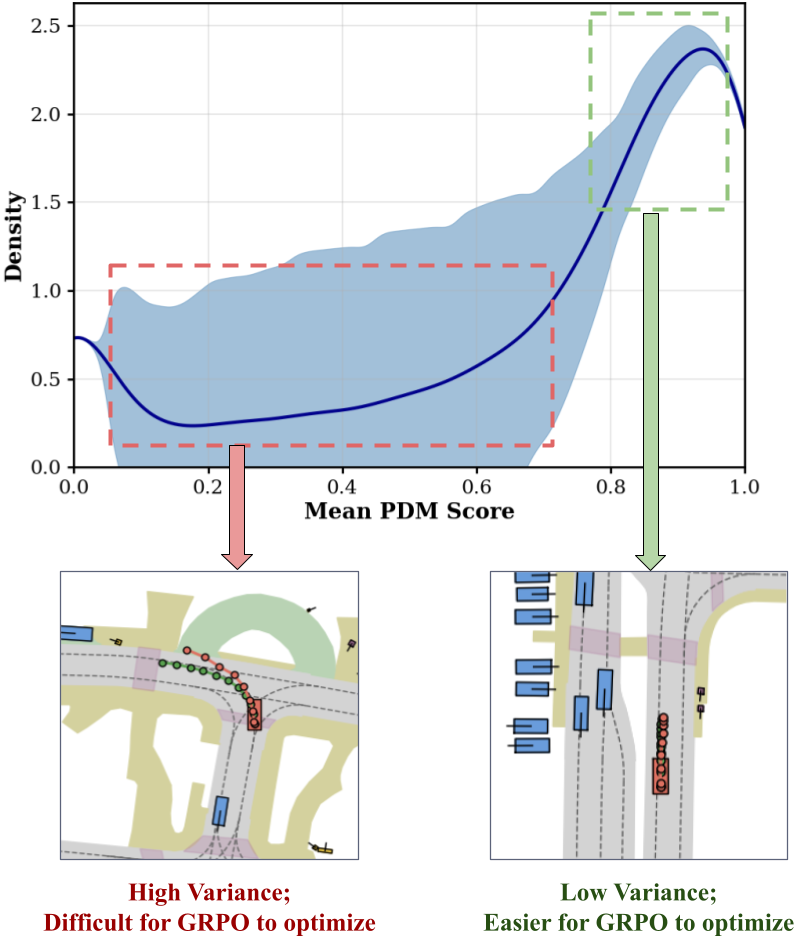}

\caption{\textbf{Reward distribution in the weak SFT model.} The group-mean PDM score is shown with band representing the mean of the corresponding group standard deviation for  \modelname{\textsc{-base}}. GRPO struggles to optimize high-variance regions (the majority) and is effective only in low-variance regions (the trajectories in green and red are for ground truth and \modelname{\textsc{-base}} prediction).} 
   \label{fig:difficulty_plot}
\end{figure}
%https://docs.google.com/presentation/d/1LguR4dDxCcsLMiVxF1emRECRNQ4RVHYPpoNQO01o2NY/edit?slide=id.g39937671229_0_6#slide=id.g39937671229_0_6

\begin{figure*}
    \centering
    \begin{subfigure}[b]{0.40\linewidth}
        \centering
        \includegraphics[width=\linewidth]{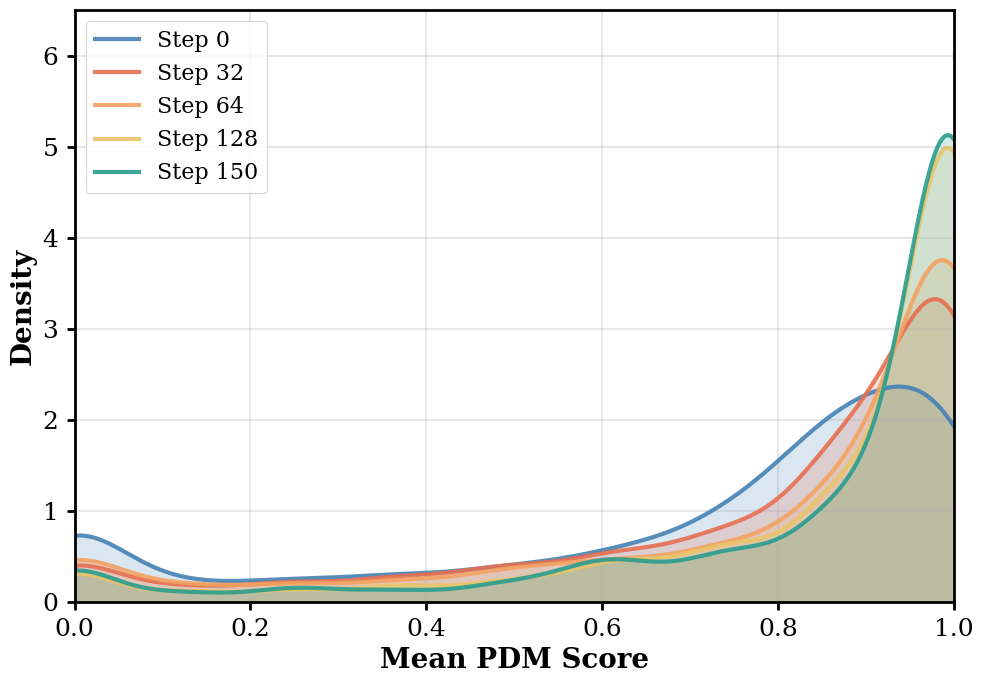}
        \caption{GRPO}
        \label{fig:grpo_steps}
    \end{subfigure}
    \hfill
    \begin{subfigure}[b]{0.40\linewidth}
        \centering
        \includegraphics[width=\linewidth]{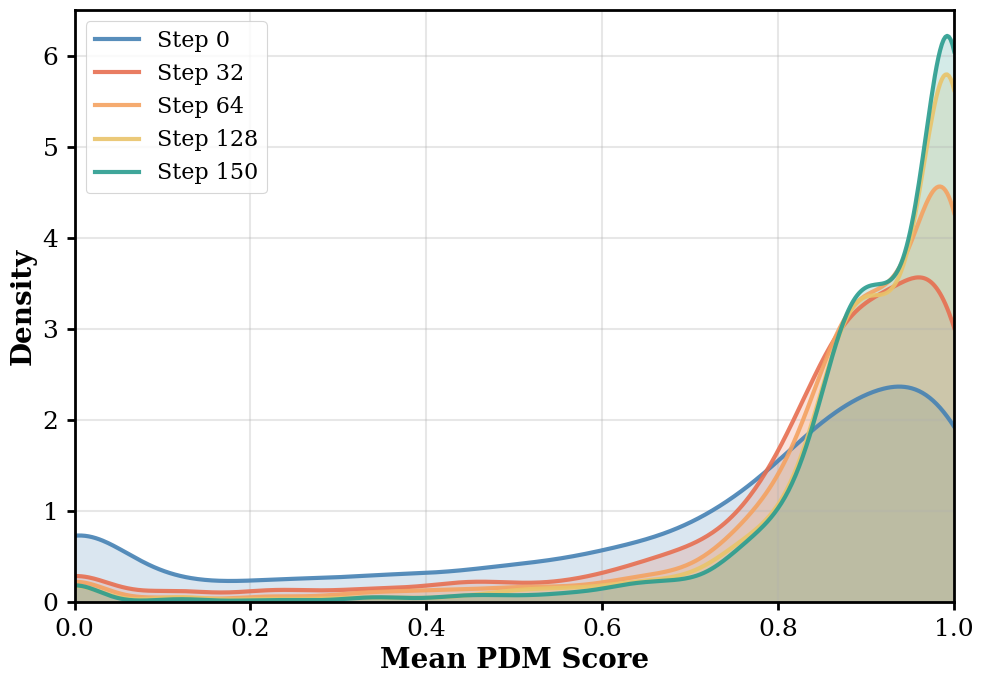}
        \caption{Dr.~GRPO}
        \label{fig:drgrpo_steps}
    \end{subfigure}
    \caption{\textbf{Evolution of group-mean PDM score during RL fine-tuning.} (a) GRPO struggles to optimize samples with high group variance during training, particularly in the range $[0.2$–$0.65]$. (b) Dr.~GRPO effectively optimizes high-variance samples during training, resulting in significant overall performance gains.}
    \label{fig:grpo_drgrpo_steps}
\end{figure*}

\begin{figure}
  \centering
   \includegraphics[width=1.0\linewidth]{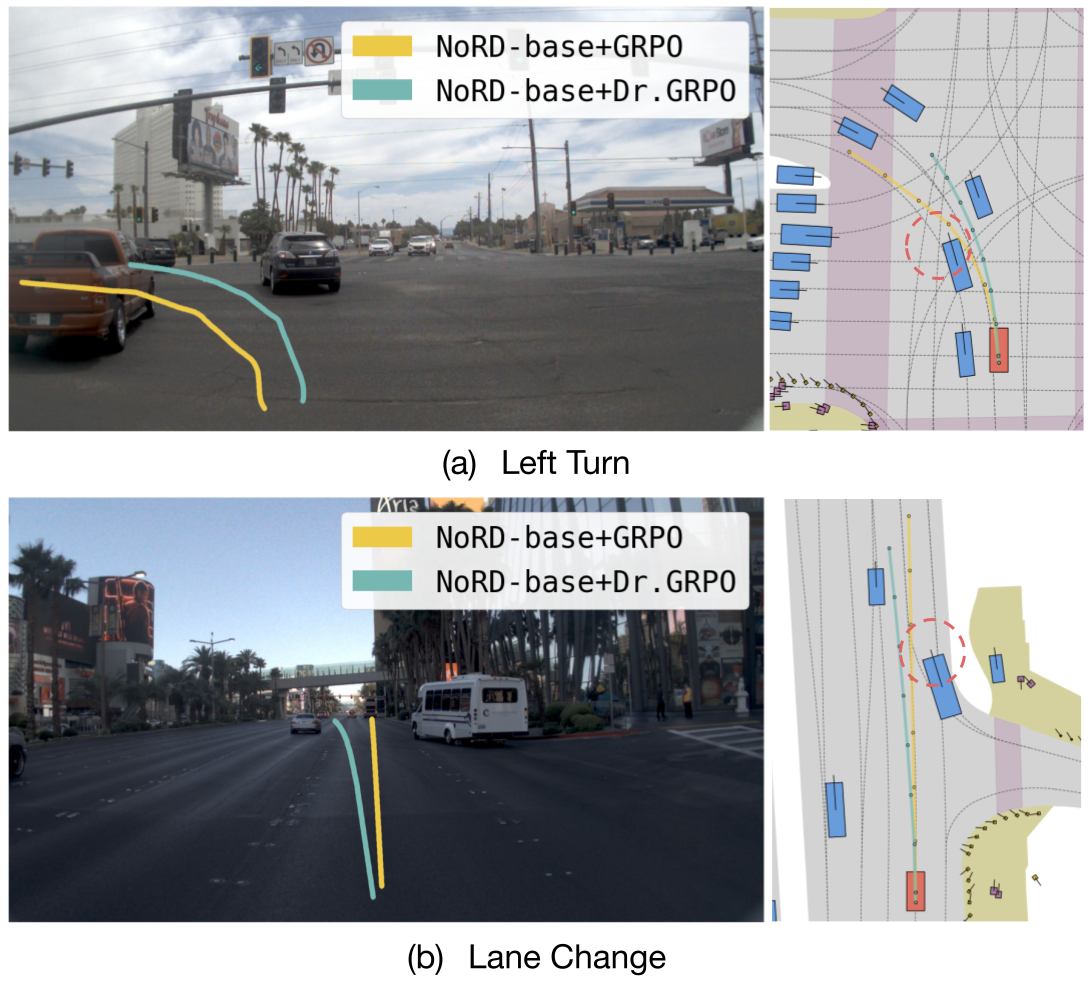}

   \caption{\textbf{Qualitative comparison of RL fine-tuning (RLFT) on the weak SFT model using GRPO and Dr.~GRPO.} With Dr.~GRPO, \modelname successfully learns complex maneuvers such as sharp turns and lane changes without collisions, whereas GRPO fails to optimize the weak SFT model (\modelname{\textsc{-base}}) and collides (in red).}
   \label{fig:comparison}
\end{figure}

\begin{table}
\caption{Comparison of RL fine-tuning (RLFT) on \modelname{\textsc{-base}} with GRPO and Dr.~GRPO on NAVSIM test set. While GRPO fails to improve \modelname{\textsc{-base}}, we get significant gains with Dr.~GRPO.}
\label{tab:grpo_drgrpo_comparison}
\centering
\begin{tabular}{@{}ll@{}}
    \toprule
    \textbf{Model} & \textbf{PDMS} $\uparrow$ \\
    \midrule
    \modelname{\textsc{-base}} & 76.66 \\
    \modelname{\textsc{-base}} + GRPO & 77.18 (+0.67\%) \\
    \modelname{\textsc{-base}} + Dr.~GRPO & 85.62 (+11.68\%) \\
    \bottomrule
 \end{tabular}
\end{table}

VLAs for autonomous driving have achieved competitive performance through a two-stage training pipeline - first SFT followed by RL post-training. This paradigm relies on large-scale domain-specific datasets that are additionally annotated with reasoning data. During RL post-training, GRPO optimizes the SFT model (\ie, the policy) for high-level objectives such as preference alignment or safety by maximizing the group-relative advantage.

However, this existing SFT-heavy approach is costly and inefficient. First, collecting and labeling  thousands or even millions of driving scenarios is resource-intensive. Second, generating reasoning traces from a teacher model increases token load, training time, and compute requirements. Finally, reasoning tokens during inference add latency, limiting real-time deployment. These challenges raise our central question: \emph{can VLAs fine-tuned on small-scale driving data without reasoning supervision achieve competitive performance, or is RL post-training inherently limited in optimizing weaker, data-efficient VLAs?}

To investigate this, we train \modelname{\textsc{-base}}, a VLA built on Qwen-2.5VL-3B-Instruct~\cite{bai2025qwen2}, using supervised fine-tuning on only 80,000 NAVSIM training samples and without reasoning annotations. \modelname{\textsc{-base}} predicts physical trajectory tokens from current images and historical vehicle states, followed by GRPO optimization on the PDM score (details in \cref{sec:nord}). The PDM score evaluates predicted trajectories in simulation across metrics such as safety, comfort, collision avoidance, and adherence to driving areas, with higher scores indicating better performance.

As shown in \cref{tab:grpo_drgrpo_comparison}, GRPO post-training leads to only a 0.67\% improvement, resulting in negligible overall gains. This outcome is inconsistent with our goal of shifting the primary learning burden from SFT to RL post-training. This minimal improvement is in stark contrast to prior works like AutoVLA~\cite{zhou2025autovla}, which have demonstrated a 9\% performance boost by post-training an SFT model, but one that was trained on 212,000+ samples and with reasoning data. This discrepancy between GRPO's effectiveness on strong versus weak SFT policies motivates a deeper investigation into the underlying cause.

% To understand this discrepancy, we analyze the reward characteristics of the training set for \modelname{\textsc{-base}}, shown in \cref{fig:difficulty_plot}. For each training example, we perform 8 rollouts and plot the distribution of group-mean PDM scores, with bands indicating the corresponding group standard deviation, averaged across groups. Our observations are as follows:
% \begin{enumerate}
% \item \textbf{High variance occurs in samples with group-mean in the range $[0.2,,0.65]$.} The PDM score penalizes collisions, off-road behavior, and other safety violations. Given the weakness of the SFT model, complex maneuvers, such as sharp turns, fail more often than they succeed, yielding low mean rewards and high variance within the rollout group.
% \item \textbf{Low variance occurs in samples with high group mean ($\geq 0.8$ and $\leq0.15$).} The weak SFT model performs almost always reliably on simple behaviors such as maintaining a straight trajectory at constant speed. These scenarios are inherently easier, resulting in consistently high reward and low intra-group variance. On the other extreme, for extremely difficult scenarios, for example out-of-distribution driving in mountains, our weak SFT model generally predicts trajectories with trivial PDM score, explaining low mean and high variance. 
% \end{enumerate}
To understand this discrepancy, we analyze the reward characteristics of the training set for \modelname{\textsc{-base}}, shown in \cref{fig:difficulty_plot}. For each training example, we perform {8 rollouts} and plot the distribution of {group-mean PDM scores}, with bands indicating the corresponding {group standard deviation}, averaged across groups. Our key observations are:

\begin{enumerate}
    \item \textbf{Low variance occurs in samples with high or very low group mean ($\geq 0.8$ or $\leq 0.15$).} The weak SFT model performs reliably on simple behaviors, such as maintaining a straight trajectory at constant speed, resulting in high mean reward and low intra-group variance. Conversely, in extremely difficult scenarios, \eg, out-of-distribution driving, the model predicts trajectories with {trivially low PDM scores}, yielding {low mean and low variance within the rollout group}. Notably, the proportion of such samples is very small, suggesting that \modelname is sufficiently expressive.
    \item \textbf{High variance occurs in samples with intermediate group-mean values in the range $[0.2, 0.65]$.} The PDM score penalizes collisions, off-road behavior, and other safety violations. Given the weakness of the SFT model, {complex maneuvers} such as sharp turns fail more often than they succeed, producing {low mean rewards and high variance} within the rollout group.    
    % \item \textbf{Low variance occurs in samples with high or very low group mean ($\geq 0.8$ or $\leq 0.15$).} The weak SFT model performs reliably on simple behaviors, such as maintaining a straight trajectory at constant speed, resulting in high mean reward and low intra-group variance. Conversely, in extremely difficult scenarios, \eg, out-of-distribution driving in mountainous regions, the model predicts trajectories with {trivially low PDM scores}, yielding {low mean and low variance within the rollout group}. Interestingly. the proportion of such samples is relatively very low suggesting that \modelname is sufficiently expressive.
\end{enumerate}
With these observations, we analyze the evolution of group-mean reward distributions across GRPO training steps (see \cref{fig:grpo_steps}). We find that the density in the high-variance region ($[0.2, 0.65]$) remains largely unchanged throughout training, whereas the density in the lowest-variance region (close to ~1) steadily increases. This pattern explains the marginal improvement in the final PDM score after GRPO post-training: GRPO primarily optimizes the small subset of samples with low intra-group reward variance while failing to improve the majority of samples with high intra-group variance. 

Our findings suggest that GRPO is fundamentally ineffective at learning from samples with high intra-group variance, which dominate the training dataset for our weak SFT model, \modelname{\textsc{-base}}, and therefore provides limited benefit for RL post-training. We interpret this failure as a form of \textit{difficulty bias} in GRPO. Originally, difficulty bias was proposed for binary reward settings, measured as the mean group reward~\cite{li2025disco}, and used to post-train LLMs for mathematical reasoning~\cite{liu2025understanding}. Building on this, our analysis shows that the limitation of optimizing weak SFT policies with GRPO for data-efficient VLA training stems from this inherent difficulty bias. To address this, we post-train \modelname{\textsc{-base}} using Dr.~GRPO, a GRPO variant originally designed to mitigate difficulty bias in LLM reasoning. Dr.~GRPO enables training with significantly less data and without any reasoning annotations, as explained in the next section.

% Formally, we can define the difficulty $\rho$ of a sample $x_i \in \mathcal{X}$ with respect to the policy $\pi: \mathcal{X} \to \mathcal{O}$ as its expected reward,  
% \[
% \rho(x_i, \pi) \coloneqq \mathbb{E}_{o \sim \pi(\cdot \mid x_i)} \big[ r(o \mid x_i) \big].
% \]
% In our case, difficulty is defined with respect to the weak SFT policy $\pi = \modelname{\textsc{-base}}$ before any RL training begins.
% This formalism directly captures our empirical findings. Our analysis \shubh{figure difficulty distribution} that the rewards are highly polarized is equivalent to stating that the initial difficulty distribution is bimodal, with nearly all of its mass clustered at the extremes ($\rho \approx 0$ and $\rho \approx 1$) and almost no mass in the ``moderately difficult" range.

% This explicitly frames our empirical finding of low-diversity rewards as a problem of an ill-posed initial difficulty distribution for RL post-training. While prior works analyze difficulty to improve the efficiency of RL on strong reasoning models, our work is the first to identify this bias as a fundamental barrier preventing weak, data-efficient VLAs from learning via RL. 

% \shubh{Reasoning-free SFT baseline}

% \shubh{Why vanilla GRPO fails?}

% This analysis shows that vanilla GRPO is fundamentally unequipped to optimize for autonomous driving benchmark metrics. The CoT-then-RL methods unintentionally bypass this problem: their CoT-SFT stage produces a policy that is already high-performing and so the subsequent GRPO stage is merely polishing a good policy.

\section{\modelname: No Reasoning for Driving}
\label{sec:nord}

\begin{figure}[t]
  \centering
   \includegraphics[width=0.9\linewidth]{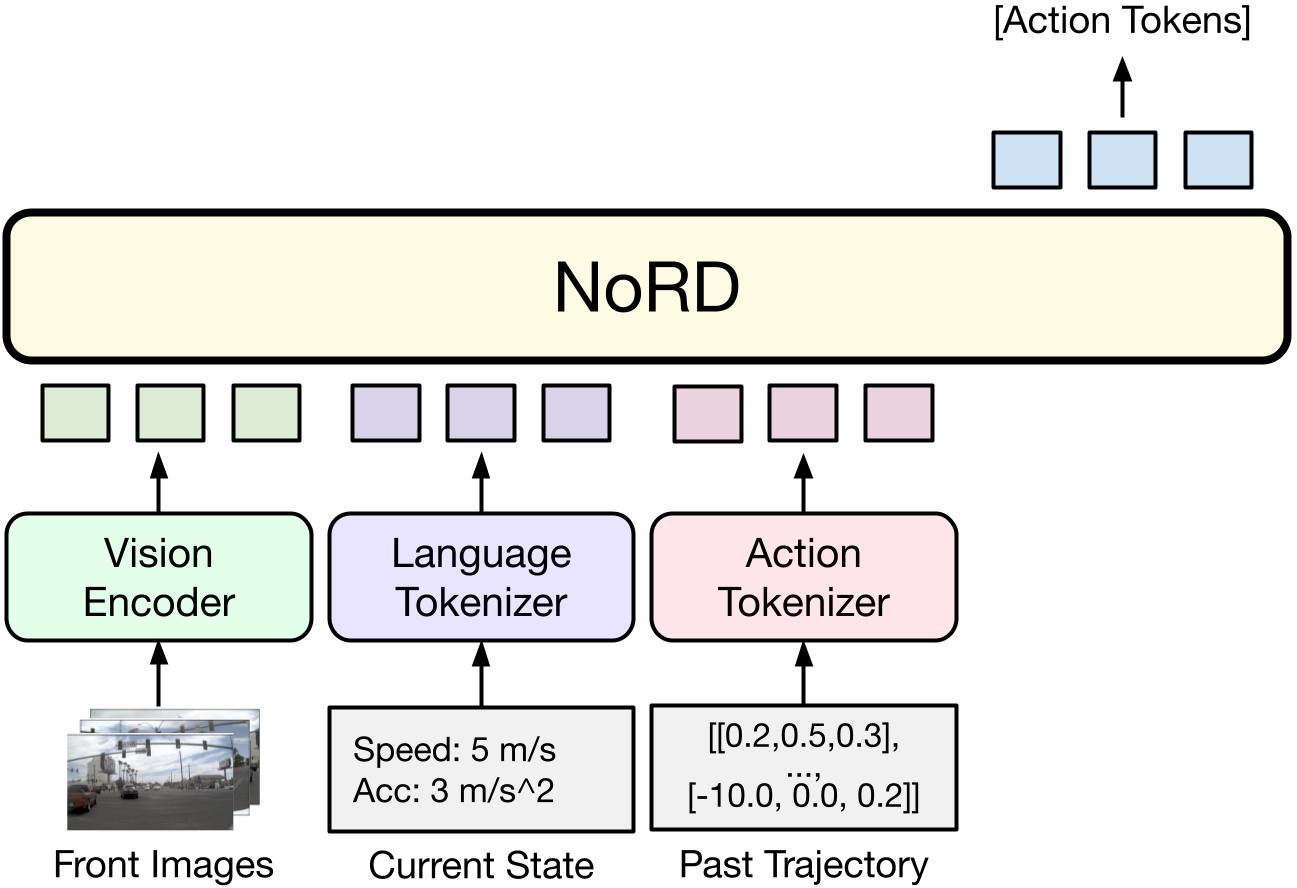}
    \caption{\textbf{Model architecture of \modelname.} \modelname directly predicts action tokens without requiring reasoning traces, enabling a significantly more efficient training and inference pipeline.}
   \label{fig:nord_arch}
\end{figure}

\modelname (\textbf{No} \textbf{R}easoning for \textbf{D}riving) is our Vision-Language-Action (VLA) model for autonomous driving, built upon {Qwen-2.5VL-3B-Instruct}. \modelname achieves both token and data efficiency by omitting reasoning annotations entirely from the training and inference stages, by emphasizing learning during the RL post-training phase rather than during supervised fine-tuning. However, as discussed in \cref{sec:difficulty_bias}, naively reducing the amount of SFT data significantly degrades performance because GRPO is ineffective in learning from samples with high intra-group variance. This section presents the design of \modelname and approach for training it effectively with limited data.

The inputs to \modelname are the past ego-trajectory, current speed, acceleration, and RGB images from the front, front-left, and front-right cameras, as shown in \cref{fig:nord_arch}. The model predicts the future ego-trajectory at 10 Hz. To improve token efficiency, we represent trajectories using k-disc tokenization~\cite{philiontrajeglish} with a vocabulary size of 2048. Specifically, all future trajectories in the training set are first interpolated to 10 Hz and segmented into 0.5 second intervals. These segments are then clustered into 2048 clusters based on the contour distance between trajectory segments. The resulting cluster centers form a discrete trajectory codebook that can reconstruct any trajectory using vocabulary tokens. These trajectory tokens  are appended to the original vocabulary of the base Qwen model initialized by sample from a multivariate normal distribution parameterized by the mean and covariance of the existing token embeddings~\cite{hewitt2021initializing}.  The model is trained in two stages: (1) \textit{Supervised Fine-Tuning} with limited data, followed by (2) \textit{RL Post-Training} using Dr.~GRPO for effective policy optimization starting from a weak SFT model, as illustrated in \cref{fig:nord_intro}.

\subsection{Supervised Fine-Tuning with Limited Data}

\modelname{\textsc{-base}} is intentionally trained on a limited dataset during supervised fine-tuning to offload the majority of learning to the subsequent RL post-training phase. We model trajectory prediction as a next-token prediction problem, where the model outputs trajectory tokens conditioned on the inputs. As expected, the reduced training data results in lower initial performance for \modelname{\textsc{-base}} (see \cref{tab:grpo_drgrpo_comparison}). In the following section, we describe how to effectively optimize this weak SFT policy using Dr.~GRPO by explicitly accounting for intra-group reward variance.

\subsection{RL Post-Training for Weak SFT Policy}
As discussed in \cref{sec:difficulty_bias}, weak SFT policies cannot be effectively optimized using standard GRPO, which can be understood as an instance of the difficulty bias problem. To address this, we employ Dr.~GRPO, a recently proposed RL fine-tuning algorithm, for post-training our weak SFT model, \modelname{\textsc{-base}}. In the original GRPO formulation, the group relative advantage is computed as
\[
\hat{A}_{i,t}^{\text{GRPO}} \coloneqq  \frac{r(o_i \mid x) - \frac{1}{G}\sum_{j=1}^G r(o_j \mid x)} {\mathrm{std}_{j=1,\ldots,G}(r(o_j \mid x))}.
\]
Here, $r(o_i \mid x)$ is the reward for sample $i$ given input $x$, $G$ is the group size, and $\mathrm{std}$ denotes the standard deviation across the group.
% Recent studies have shown that this formulation unintentionally favors groups with low variance. When the group standard deviation is small (e.g., $<1$), the group-relative advantage becomes disproportionately large, whereas for high-variance groups it becomes heavily attenuated. In our setting, this causes GRPO to be ineffective at optimizing the weak SFT policy; the dataset contains a substantially larger proportion of high-variance groups, which typically correspond to more complex maneuvers, while GRPO concentrates its updates on the small set of near-perfect or to the very complex out-of-distribution scenes, low-variance cases.
Recent studies have shown that this formulation unintentionally favors groups with low reward variance~\cite{liu2025understanding}. When the standard deviation of the reward within the group  is small (\ie., $<<1$), the group-relative advantage is disproportionately large, whereas it is heavily attenuated for groups with high reward variance. 
This poses a major problem for us since \modelname{\textsc{-base}}, being a weak SFT model, produces groups with high intra-group variance during the GRPO rollout for the majority of samples ( \cref{fig:difficulty_plot}).
% , corresponding to complex maneuvers such as turns or driving in traffic. In our setting, this causes GRPO to be ineffective at optimizing the weak SFT policy. In contrast, the low-variance groups consist of either very easy scenarios, \eg, driving straight at constant speed, which consistently yield near-perfect rewards, or extremely difficult out-of-distribution cases, which consistently yield trivial rewards. GRPO disproportionately focuses on these variance groups, leaving the majority of complex, high-variance maneuvers under-optimized.
Dr.~GRPO mitigates difficulty bias by removing the standard deviation term from the group relative advantage, enabling more effective optimization of weak policies. Notably, while other variants like VD-GRPO~\cite{tang2026planr1} preserve absolute reward magnitudes to balance objective priorities (e.g., safety vs. comfort), Dr.~GRPO ensures that 'hard' scenarios contribute a sufficient gradient signal. Additionally, we employ DAPO-style asymmetric clipping to prevent entropy collapse during RL training and follow \citet{liu2025understanding} by not using KL-divergence regularization. The resulting Dr.~GRPO post-training objective is given by:

\begin{align}
\hat{A}_{i,t}^{\text{DrGRPO}} &= 
r(o_i \mid x) - \frac{1}{G}\sum_{j=1}^G r(o_i \mid x), \\
L_{\text{DrGRPO}} &= \sum_{t=1}^{|o_i|} 
\min\Bigg(
\frac{\pi_\theta(o_{i,t}|q,o_{i,<t})}
     {\pi_{\theta_\text{old}}(o_{i,t}|q,o_{i,<t})} \hat{A}_{i,t}^{\text{DrGRPO}}, \notag\\
&\quad \text{clip}\Big(
\frac{\pi_\theta(o_{i,t}|q,o_{i,<t})}
     {\pi_{\theta_\text{old}}(o_{i,t}|q,o_{i,<t})}, 1-\epsilon_\text{l}, 1+\epsilon_\text{h}
\Big) \hat{A}_{i,t}^{\text{DrGRPO}}
\Bigg)
\end{align}

This formulation enables \modelname, \ie, \modelname{\textsc{-base}} trained with Dr.~GRPO, to achieve improved performance during RL post-training by mitigating difficulty bias and stabilizing policy optimization for data- and token-efficient VLAs. We find that with Dr.~GRPO finetuning, \modelname{\textsc{-base}} learns from mid-variance samples, leading to an overall improvement of 11.68\% from the base model (as compared to 0.67\% with GRPO). We notice that Dr.~GRPO is able to optimize even on the samples with high intra-group variance, as shown in \cref{fig:drgrpo_steps}. This enables \modelname even learn complex maneuvers, as compared to the GRPO counterpart (see \cref{fig:comparison}).

\label{sec:nord}
\section{Experiments}
\subsection{Datasets}
\textbf{NAVSIM}~\cite{Dauner2024NEURIPS}: NAVSIM is a curated redistribution of the OpenScenes dataset, comprising real-world urban driving scenarios. The dataset comprises 120 hours of driving data from OpenScene. The dataset features diverse and challenging traffic situations, providing synchronized $360^{\circ}$ camera imagery, LiDAR scans, HD map data, and bounding-box annotations of dynamic agents, along with historical control signals. The task is to predict the ego-vehicle trajectory for the next 4 seconds at 2~Hz, with performance evaluated by executing the predicted trajectory within a simulation environment, scored using PDM Score in terms of driving safety, progress and comfort metrics.

\textbf{Waymo Vision-Based End-to-End Dataset (WaymoE2E)}~\cite{waymo2025e2e}: WaymoE2E is a challenging long-tail dataset providing $360^{\circ}$ camera views and ego-vehicle trajectories. 
Validation and test scenarios include three alternative trajectories labeled with human per scene, representing varying driving preferences and scored in the range $[4, 10]$. 
Predicted trajectories are evaluated using the Rated Feedback Score (RFS), which measures weighted similarity with the reference preference trajectories. WaymoE2E contains 4,021 challenging driving segments, partitioned into 2,037 training, 479 validation, and 1,505 test segments. 

\subsection{Implementation Details}
We use {Qwen-2.5VL-3B-Instruct} as our base model, as it offers a good trade-off between model capacity and computational efficiency. The model is fine-tuned on the NAVSIM and WaymoE2E datasets separately using 16 A100 GPUs with a batch size of 128. We employ the AdamW optimizer with a learning rate of $5\times10^{-5}$ and a cosine decay schedule, fine-tuning all layers of {Qwen-2.5VL-3B}. This stage yields the \modelname{\textsc{-base}} model. Subsequently, we apply Dr.~GRPO for RL post-training. For NAVSIM, we use 30 A100 GPUs to optimize the base model for 160 steps with a constant learning rate of $5\times10^{-6}$. For WaymoE2E, we post-train the model for 150 steps on 32 GPUs with a learning rate of $1\times10^{-6}$.

The RL post-training pipeline is implemented in \texttt{verl}~\cite{sheng2024verl} with Fully Sharded Data Parallel (FSDP) for memory-efficient training and \texttt{vLLM}~\cite{kwon2023vllm} for rollout generation. During rollouts, we set the group size to 8 and fix the sampling temperature to 1.0. For validation, we employ deterministic sampling with a temperature of 0.01. For NAVSIM, we use the PDM score as the primary reward function, while for WaymoE2E, we employ the normalized RFS score. In both cases, we include additional rewards for trajectory length and output format, each weighted by 0.25, and then normalize to $[0,1]$.

\subsection{Results}

\begin{table}
\caption{Test results on the Waymo Vision-based End-to-End Driving Benchmark. \modelname achieves competitive performance, without reasoning or ensembling.}
\label{tab:waymo_results}
\centering
\begin{tabular}{@{}lcccc@{}}
\toprule
\textbf{Model} & \makecell{\textbf{w/o} \\ \textbf{Reason}} & \makecell{\textbf{w/o} \\ \textbf{Ensemble}} & \textbf{RFS}$\uparrow$ & \textbf{ADE@3}$\downarrow$ \\
\midrule
Poutine~\cite{rowe2025poutine} & \bad & \good & {7.986} & 1.2055 \\
HMVLM~\cite{wang2025hmvlm} & \bad & \good & 7.736 & 1.3269 \\
DiffusionLTF & \good & \bad & 7.717 & 1.3561 \\
UniPlan & \good & \bad &  7.692 & 1.3083 \\
AutoVLA~\cite{zhou2025autovla} & \bad & \good &  7.556 & 1.3507 \\
\midrule
\textbf{\modelname} & \good & \good &  7.709 & 1.2504 \\
\bottomrule
 \end{tabular}
\end{table}

% \begin{table}
% \caption{Training dataset size comparison for Waymo End-to-End Driving Challenge and NAVSIM benchmark. Albeit the smallest training dataset size, \modelname achieves competitive performance. (For fair comparison, VLA models have been underlined).}
% \label{tab:dataset_size}
% \centering
% \begin{tabular}{@{}lcr@{}}
% \toprule
% Benchmark & Model & Dataset Size $\downarrow$ \\
% \midrule
% \textit{Waymo E2E} & & \\
% & \underline{AutoVLA} & 264,800\\
% & \underline{Poutine} & 240,500  \\
% & \underline{UniPlan} & 135,000 \\
% & DiffusionLTF & 94,500\\
% & \underline{HMVLM} & 63,000\\
% & \underline{\textbf{\modelname}} & 12,600 \\
% \midrule
% \textit{NAVSIM} & & \\
% & \underline{RecogDrive} &  3,100,000  \\
% & \underline{AutoVLA} & 233,800 \\
% & DiffusionDrive &  131,000 \\
% & UniAD & 103,000  \\
% & Transfuser & 103,000  \\
% & Hydra-MDP &  103,000 \\
% & \underline{\textbf{\modelname}} & 88,500 \\
% \bottomrule
%  \end{tabular}
% \end{table}

\begin{table*}
\caption{Test results on NAVSIM benchmark (navtest subset). \modelname achieves competitive performance (\textit{w/o R}: Without reasoning data, \textit{w/o L}: without LiDAR data, and \textit{C}: Number of RGB frames; *~\textit{BoN} refers to the average over best score per sample out of 6 outputs with different random seeds).}
\label{tab:navsim}
\centering
\begin{tabular}{@{}lcccccccccc@{}}
\toprule
% Method & w/o R & w/o L & C & PDMS$\uparrow$ & Collision$\uparrow$  & DAC$\uparrow$  & Direction$\uparrow$  & Progress$\uparrow$  & TTC$\uparrow$  & Comfort$\uparrow$ \\
\textbf{Method} & \makecell{\textbf{w/o} \\ \textbf{R}} & \makecell{\textbf{w/o} \\ \textbf{L}} & \textbf{C} & \textbf{PDMS}$\uparrow$ & \textbf{Collision}$\uparrow$  & \textbf{DAC}$\uparrow$  & \textbf{Direction}$\uparrow$  & \textbf{Progress}$\uparrow$  & \textbf{TTC}$\uparrow$  & \textbf{Comfort}$\uparrow$ \\
\midrule
\multicolumn{11}{@{}l}{\textit{BEV-based Methods}} \\[5pt]
UniAD~\cite{hu2023planning} & \good & \good & 32 & 83.4 & 97.7 & 91.9 & - & 78.8 & 92.9 & \textbf{100} \\
Transfuser~\cite{chitta2022transfuser} & \good & \bad & 3 & 84.0 & 97.7 & 92.8 &  \textbf{97.9} & 79.2 & 92.8 & \textbf{100} \\
Hydra-MDP~\cite{li2024hydra} & \good & \bad & 3 & 86.5 & 98.2 & 96.2 &  95.8 & 78.7 & 94.6 & \textbf{100} \\
DiffusionDrive~\cite{liao2025diffusiondrive} & \good & \bad & 3 & 88.1 & 98.2 & 96.2 &  - & 82.2 & 94.7 & 88.1 \\
\midrule
\multicolumn{11}{@{}l}{\textit{VLA-based Methods}} \\[5pt]
AutoVLA~\cite{zhou2025autovla} & \bad & \good & 12 & 89.1 & 98.4 & 95.6 & 95.4 & 81.9 & \textbf{98.0} & 99.9 \\
AutoVLA-BoN*~\cite{zhou2025autovla} & \bad & \good & 12 & 92.1 & 99.1 & 97.1 & 95.5 & \textbf{87.6} & 97.1 & \textbf{100} \\
RecogDrive~\cite{li2025recogdrive} & \bad & \good & 12 & 89.6 & 98.2 & 97.9 & - & 83.5 & 95.2 & 99.8 \\
\midrule
\textbf{\modelname} & \good & \good & 3 & 85.6 & 97.6 & 94.9 & 95.9 & 79.3 & 93.5 & \textbf{100} \\
\textbf{\modelname-BoN*} & \good & \good & 3 & \textbf{92.4} & \textbf{99.2} & \textbf{98.3} & 95.9 & 86.4 & 97.8 & 99.9 \\
\bottomrule
\end{tabular}
\end{table*}

% \begin{figure}[t]
%     \centering
%     \includegraphics[width=\linewidth]{navsim_pareto.pdf}
%     \caption{\textbf{Pareto-optimal curve for NAVSIM.} \modelname is the only VLA in the high-performance, high data-efficiency region, relying solely on RGB camera inputs for end-to-end driving.}
%     \label{fig:navsim_pareto}
% \end{figure}

% \begin{figure}[t]
%     \centering
%     \includegraphics[width=\linewidth]{waymo_pareto.pdf}
%     \caption{\textbf{Pareto-optimal curve for WaymoE2E.} \modelname is the only VLA that achieves a competitive RFS score with a fraction of the training data, while requiring neither ensembling nor reasoning supervision.}
%     \label{fig:waymo_pareto}
% \end{figure}

\begin{figure*}
    \centering

    \begin{subfigure}[b]{0.42\linewidth}
        \centering
        \includegraphics[width=\linewidth]{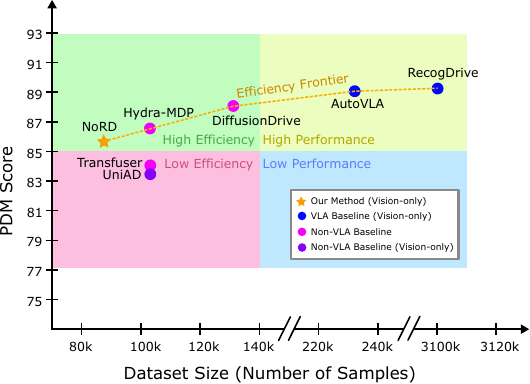}
        \caption{NAVSIM}
        \label{fig:navsim_pareto}
    \end{subfigure}
    \hfill
    \begin{subfigure}[b]{0.42\linewidth}
        \centering
        \includegraphics[width=\linewidth]{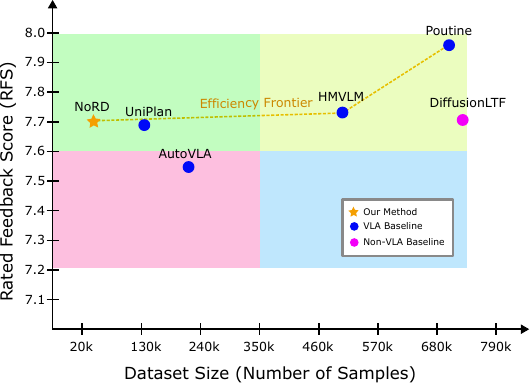}
        \caption{WaymoE2E}
        \label{fig:waymo_pareto}
    \end{subfigure}

    \caption{\textbf{Pareto-optimal curves on two driving benchmarks.}
    (a) \modelname is the only VLA in NAVSIM operating in the high-performance, high–data-efficiency region using only RGB inputs.  
    (b) \modelname achieves competitive RFS on WaymoE2E with a fraction of the training data, without ensembling or reasoning supervision. Shaded regions provide a qualitative categorization of model efficiency and performance for ease of visualization.}
    \label{fig:pareto_combined}
\end{figure*}

\begin{figure*}
    \centering
    \includegraphics[width=0.97\linewidth]{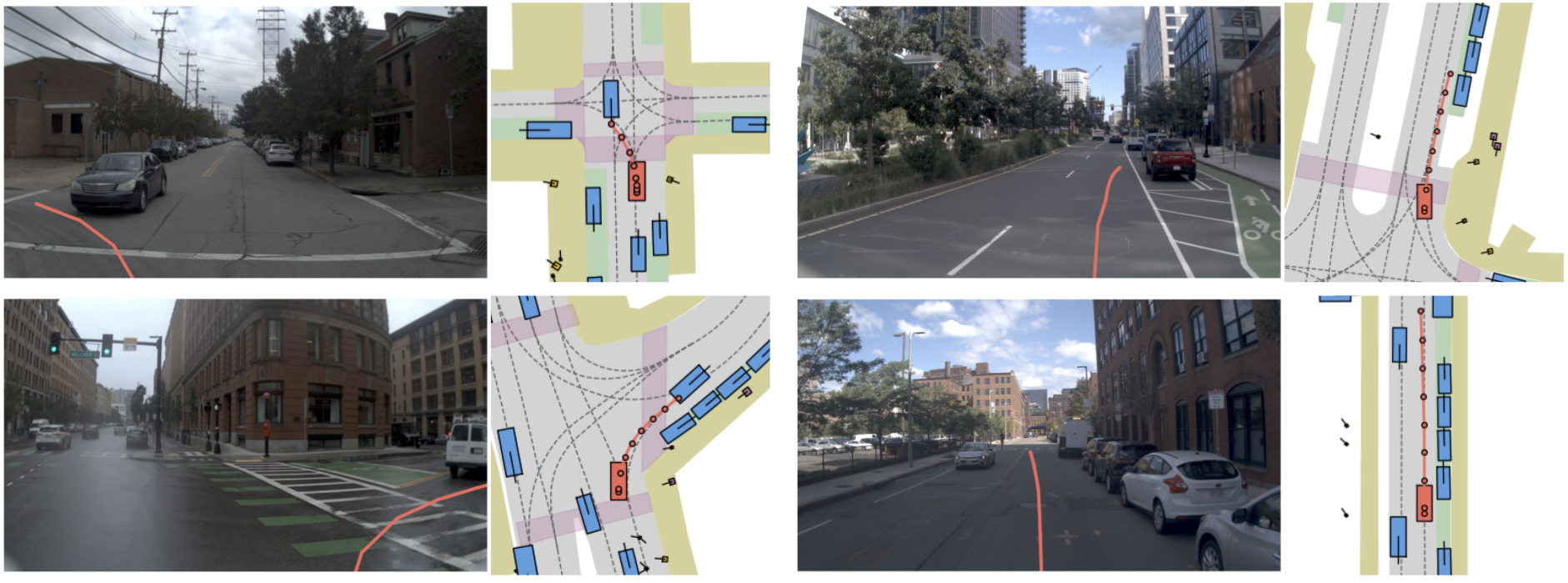}
    \caption{\textbf{Qualitative Results on NAVSIM (navtest subset)}. \modelname safely executes sharp turns, respects traffic lights, and avoids collisions, demonstrating robust driving behavior. The predicted trajectory is shown in red.}
    \label{fig:navsim_examples}
\end{figure*}

\begin{figure}
    \centering
    \includegraphics[width=0.9\linewidth]{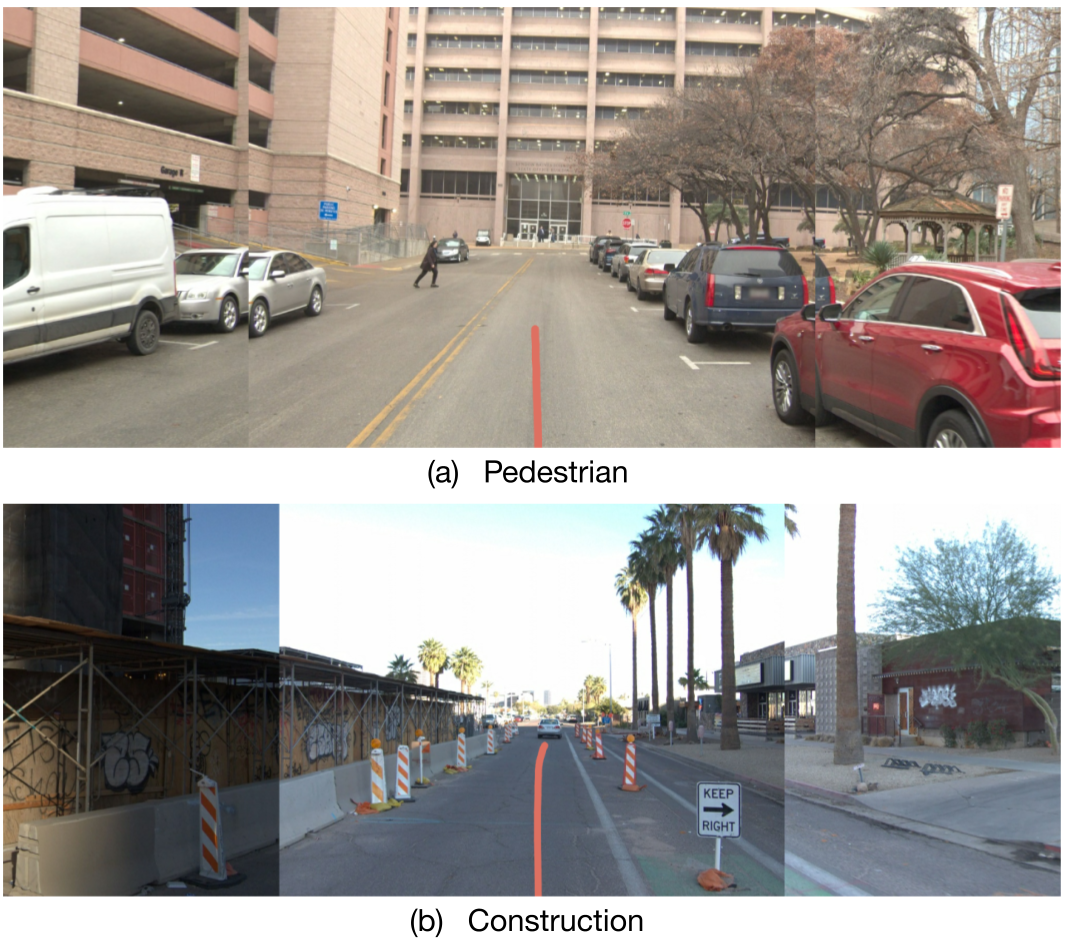}
    \caption{\textbf{Qualitative Results on WaymoE2E test set}. \modelname drives safely in challenging out-of-distribution scenarios like unsafe pedestrian crossing and construction site. The predicted trajectory is shown in red (stitched, center-cropped for visualization).}
    \label{fig:waymo_examples}
\end{figure}

\begin{figure}
        \centering
        \includegraphics[width=1\linewidth]{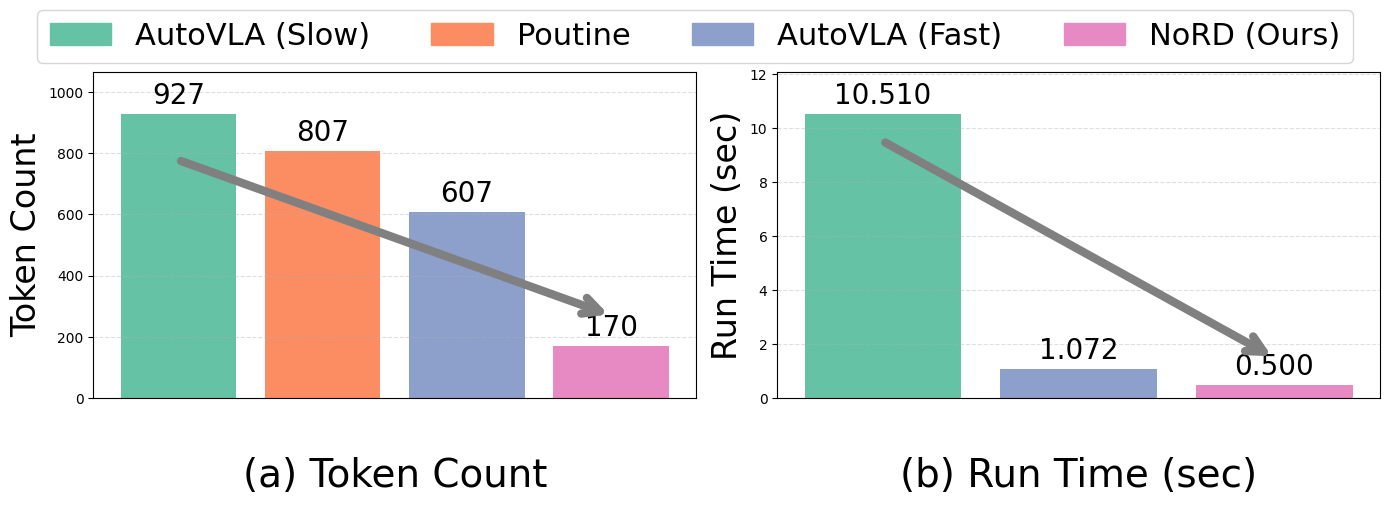}
        \caption{\textbf{Comparison of token and runtime efficiency.} \modelname is the most (a) token and (b) runtime efficient VLA.}
        \label{fig:nord_efficient}
\end{figure}

\noindent\textbf{WaymoE2E performance:} The WaymoE2E dataset poses a challenging evaluation setting that emphasizes robustness under out-of-distribution driving scenarios. Consequently, most existing approaches rely on large-scale training datasets and explicit reasoning annotations to achieve competitive performance (\cref{tab:waymo_results}). In contrast, \modelname attains a RFS of 7.709, ranking as the third best-performing VLA on the benchmark, while being the only top model trained without reasoning traces or ensembling. Remarkably, \modelname achieves this with merely 12,000 samples for supervised training and 450 samples for RLFT, whereas Poutine and HMVLM require $17\times$ and $12\times$ larger datasets for only marginal RFS gains. Furthermore, \modelname surpasses all other competitive models on the ADE metric, despite a $\geq6\times$ reduction in training data, underscoring its strong generalization ability, as shown in \cref{fig:waymo_examples}.
\newline
\newline
\noindent\textbf{NAVSIM performance:} The NAVSIM benchmark rigorously evaluates trajectory prediction by executing models in a simulator and scoring them using the PDM metric, a weighted measure of high-level driving factors such as comfort, time-to-collision, and ego-progress. As shown in \cref{tab:navsim} (with qualitative results in \cref{fig:navsim_examples}), \modelname is the only model that requires no reasoning traces, relies solely on 3 camera frames, and uses no additional features like, LiDAR and HD Map. While other VLA models, such as AutoVLA and RecogDrive, require $1.6\times$ and $34\times$ more training data, \modelname achieves competitive performance with fewer than 90,000 samples. We also evaluate the best-of-N performance,  where the oracle selects the best trajectory out of 6 predictions based on the PDM score. In this configuration, \modelname-BoN surpasses reasoning-based AutoVLA-BoN, achieving a PDM score of 92.4, highlighting its capabilities and data-efficiency. 
\newline
\newline
\noindent\textbf{Efficiency and Scalability:} A central contribution of our work is the remarkable data efficiency of \modelname, as highlighted in the Pareto-front analyses (\cref{fig:navsim_pareto} and \cref{fig:waymo_pareto}). On both benchmarks, \modelname establishes a competitive performance baseline while operating in the high-efficiency (\ie, low data) regime. While some VLA-based methods eventually achieve marginally higher absolute scores, they do so at a prohibitive data cost of at least $3\times$ more data. \modelname, in contrast, firmly establishes itself on the efficiency frontier, presenting an optimal and practical trade-off.  While maintaining data efficiency, since it directly predicts the trajectory tokens, it is extremely lightweight and this enables it to achieve significantly lower inference time and token count, as compared to other VLAs as shown in \cref{fig:nord_efficient}. Our findings strongly suggest that high-performance autonomous driving VLAs do not necessarily require large datasets, paving the way for more accessible and scalable yet efficient models.

% \begin{table}
% \caption{Effect of k-disc vocabulary size on the performance of \modelname.}
% \label{tab:vocab_size}
% \centering
% \begin{tabular}{@{}ll@{}}
%     \toprule
%     \textbf{Vocabulary Size} & \textbf{PDMS} $\uparrow$ \\
%     \midrule
%     512 & 83.07 \\
%     2048 & 85.62 \\
%     \bottomrule
%  \end{tabular}
% \end{table}
\label{sec:experiments}
\section{Conclusion}
\label{sec:conclusion}
We proposed \modelname, a reasoning-free, data-efficient VLA for autonomous driving. \modelname achieves strong performance while eliminating language reasoning and significantly reducing training data requirements. By analyzing rewards and modifying training pipelines, we demonstrate that VLAs can be trained with substantially fewer samples while improving token efficiency and inference speed. While Dr.~GRPO mitigates difficulty bias better than GRPO, it remains imperfect~\cite{li2025disco}, leaving room for future work. Importantly, \modelname does not suggest that VLAs cannot benefit from language-based reasoning; rather, it shows that efficient, high-performing VLAs can be trained without reasoning and large-scale datasets, pushing the boundaries of data and inference efficiency.

% {
%     \small
%     \bibliographystyle{ieeenat_fullname}
%     \bibliography{main}

\begin{thebibliography}{50}
\providecommand{\natexlab}[1]{#1}
\providecommand{\url}[1]{\texttt{#1}}
\expandafter\ifx\csname urlstyle\endcsname\relax
  \providecommand{\doi}[1]{doi: #1}\else
  \providecommand{\doi}{doi: \begingroup \urlstyle{rm}\Url}\fi

\bibitem[An et~al.(2025)An, Xie, Li, Li, Zhang, Gong, Zhong, Xu, Qiu, Wang, and Kong]{an2025polaris}
Chenxin An, Zhihui Xie, Xiaonan Li, Lei Li, Jun Zhang, Shansan Gong, Ming Zhong, Jingjing Xu, Xipeng Qiu, Mingxuan Wang, and Lingpeng Kong.
\newblock Polaris: A post-training recipe for scaling reinforcement learning on advanced reasoning models, 2025.

\bibitem[Bai et~al.(2025)Bai, Chen, Liu, Wang, Ge, Song, Dang, Wang, Wang, Tang, et~al.]{bai2025qwen2}
Shuai Bai, Keqin Chen, Xuejing Liu, Jialin Wang, Wenbin Ge, Sibo Song, Kai Dang, Peng Wang, Shijie Wang, Jun Tang, et~al.
\newblock Qwen2. 5-vl technical report.
\newblock \emph{arXiv preprint arXiv:2502.13923}, 2025.

\bibitem[Caesar et~al.(2020)Caesar, Bankiti, Lang, Vora, Liong, Xu, Krishnan, Pan, Baldan, and Beijbom]{caesar2020nuscenes}
Holger Caesar, Varun Bankiti, Alex~H Lang, Sourabh Vora, Venice~Erin Liong, Qiang Xu, Anush Krishnan, Yu Pan, Giancarlo Baldan, and Oscar Beijbom.
\newblock nuscenes: A multimodal dataset for autonomous driving.
\newblock In \emph{Proceedings of the IEEE/CVF conference on computer vision and pattern recognition}, pages 11621--11631, 2020.

\bibitem[Chen et~al.(2025{\natexlab{a}})Chen, Liu, Liang, Huang, Zhang, and He]{chen2025unlockingpotentialdifficultyprior}
Mingrui Chen, Haogeng Liu, Hao Liang, Huaibo Huang, Wentao Zhang, and Ran He.
\newblock Unlocking the potential of difficulty prior in rl-based multimodal reasoning, 2025{\natexlab{a}}.

\bibitem[Chen et~al.(2025{\natexlab{b}})Chen, Wang, and Zhang]{chen2025drivinggpt}
Yuntao Chen, Yuqi Wang, and Zhaoxiang Zhang.
\newblock Drivinggpt: Unifying driving world modeling and planning with multi-modal autoregressive transformers.
\newblock In \emph{Proceedings of the IEEE/CVF International Conference on Computer Vision (ICCV)}, pages 26890--26900, 2025{\natexlab{b}}.

\bibitem[Chi et~al.(2025)Chi, Gao, Liu, Liu, Liu, Li, Yang, Yu, Wang, Li, et~al.]{chi2025impromptu}
Haohan Chi, Huan-ang Gao, Ziming Liu, Jianing Liu, Chenyu Liu, Jinwei Li, Kaisen Yang, Yangcheng Yu, Zeda Wang, Wenyi Li, et~al.
\newblock Impromptu vla: Open weights and open data for driving vision-language-action models.
\newblock \emph{arXiv preprint arXiv:2505.23757}, 2025.

\bibitem[Chitta et~al.(2022)Chitta, Prakash, Jaeger, Yu, Renz, and Geiger]{chitta2022transfuser}
Kashyap Chitta, Aditya Prakash, Bernhard Jaeger, Zehao Yu, Katrin Renz, and Andreas Geiger.
\newblock Transfuser: Imitation with transformer-based sensor fusion for autonomous driving.
\newblock \emph{IEEE transactions on pattern analysis and machine intelligence}, 45\penalty0 (11):\penalty0 12878--12895, 2022.

\bibitem[Chu et~al.(2025)Chu, Huang, Zhang, Wei, and Wang]{chu2025gpg}
Xiangxiang Chu, Hailang Huang, Xiao Zhang, Fei Wei, and Yong Wang.
\newblock Gpg: A simple and strong reinforcement learning baseline for model reasoning.
\newblock \emph{arXiv preprint arXiv:2504.02546}, 2025.

\bibitem[Corbière et~al.(2025)Corbière, Roburin, Montariol, Bosselut, and Alahi]{drivingvqa2025}
Charles Corbière, Simon Roburin, Syrielle Montariol, Antoine Bosselut, and Alexandre Alahi.
\newblock Retrieval-based interleaved visual chain-of-thought in real-world driving scenarios, 2025.

\bibitem[Cui et~al.(2025)Cui, Yuan, Wang, Wang, Li, He, Fan, Yu, Xu, Chen, et~al.]{cui2025prime}
Ganqu Cui, Lifan Yuan, Zefan Wang, Hanbin Wang, Wendi Li, Bingxiang He, Yuchen Fan, Tianyu Yu, Qixin Xu, Weize Chen, et~al.
\newblock Process reinforcement through implicit rewards.
\newblock \emph{arXiv preprint arXiv:2502.01456}, 2025.

\bibitem[Dauner et~al.(2024)Dauner, Hallgarten, Li, Weng, Huang, Yang, Li, Gilitschenski, Ivanovic, Pavone, Geiger, and Chitta]{Dauner2024NEURIPS}
Daniel Dauner, Marcel Hallgarten, Tianyu Li, Xinshuo Weng, Zhiyu Huang, Zetong Yang, Hongyang Li, Igor Gilitschenski, Boris Ivanovic, Marco Pavone, Andreas Geiger, and Kashyap Chitta.
\newblock Navsim: Data-driven non-reactive autonomous vehicle simulation and benchmarking.
\newblock In \emph{Advances in Neural Information Processing Systems (NeurIPS)}, 2024.

\bibitem[Deng et~al.(2025)Deng, Yan, Wei, Ma, Yang, Chen, Zhang, Yang, Zhang, Zhang, et~al.]{deng2025graspvla}
Shengliang Deng, Mi Yan, Songlin Wei, Haixin Ma, Yuxin Yang, Jiayi Chen, Zhiqi Zhang, Taoyu Yang, Xuheng Zhang, Wenhao Zhang, et~al.
\newblock Graspvla: a grasping foundation model pre-trained on billion-scale synthetic action data.
\newblock \emph{arXiv preprint arXiv:2505.03233}, 2025.

\bibitem[Fan et~al.(2025)Fan, Yang, Liu, Wu, Che, Liu, and Wan]{fan2025diffusion}
Shichao Fan, Quantao Yang, Yajie Liu, Kun Wu, Zhengping Che, Qingjie Liu, and Min Wan.
\newblock Diffusion trajectory-guided policy for long-horizon robot manipulation.
\newblock \emph{arXiv preprint arXiv:2502.10040}, 2025.

\bibitem[Fu et~al.(2025)Fu, Zhang, Zhao, Cui, Liang, Zhang, Zhang, Xie, Wang, and Bai]{fu2025orion}
Haoyu Fu, Diankun Zhang, Zongchuang Zhao, Jianfeng Cui, Dingkang Liang, Chong Zhang, Dingyuan Zhang, Hongwei Xie, Bing Wang, and Xiang Bai.
\newblock Orion: A holistic end-to-end autonomous driving framework by vision-language instructed action generation.
\newblock In \emph{Proceedings of the IEEE/CVF International Conference on Computer Vision}, 2025.

\bibitem[Guo et~al.(2025)Guo, Yang, Zhang, Song, Wang, Zhu, Xu, Zhang, Ma, Bi, et~al.]{guo2025deepseek}
Daya Guo, Dejian Yang, Haowei Zhang, Junxiao Song, Peiyi Wang, Qihao Zhu, Runxin Xu, Ruoyu Zhang, Shirong Ma, Xiao Bi, et~al.
\newblock Deepseek-r1 incentivizes reasoning in llms through reinforcement learning.
\newblock \emph{Nature}, 645\penalty0 (8081):\penalty0 633--638, 2025.

\bibitem[Hewitt(2021)]{hewitt2021initializing}
John Hewitt.
\newblock Initializing new word embeddings for pretrained language models, 2021.

\bibitem[Hu et~al.(2023)Hu, Yang, Chen, Li, Sima, Zhu, Chai, Du, Lin, Wang, et~al.]{hu2023planning}
Yihan Hu, Jiazhi Yang, Li Chen, Keyu Li, Chonghao Sima, Xizhou Zhu, Siqi Chai, Senyao Du, Tianwei Lin, Wenhai Wang, et~al.
\newblock Planning-oriented autonomous driving.
\newblock In \emph{Proceedings of the IEEE/CVF conference on computer vision and pattern recognition}, pages 17853--17862, 2023.

\bibitem[Hwang et~al.(2025)Hwang, Xu, Lin, Hung, Ji, Choi, Huang, He, Covington, Sapp, Zhou, Guo, Anguelov, and Tan]{hwang2025emma}
Jyh-Jing Hwang, Runsheng Xu, Hubert Lin, Wei-Chih Hung, Jingwei Ji, Kristy Choi, Di Huang, Tong He, Paul Covington, Benjamin Sapp, Yin Zhou, James Guo, Dragomir Anguelov, and Mingxing Tan.
\newblock {EMMA}: End-to-end multimodal model for autonomous driving.
\newblock \emph{Transactions on Machine Learning Research}, 2025.

\bibitem[Jia et~al.(2023)Jia, Mao, Liu, Zhao, Wen, Zhang, Zhang, and Wang]{jia2023adriver1}
Fan Jia, Weixin Mao, Yingfei Liu, Yucheng Zhao, Yuqing Wen, Chi Zhang, Xiangyu Zhang, and Tiancai Wang.
\newblock Adriver-i: A general world model for autonomous driving, 2023.

\bibitem[Jiang et~al.(2025)Jiang, Gao, Sun, Wang, Wang, Chai, Cao, Heng, Jiang, Dong, et~al.]{jiang2025diffvla}
Anqing Jiang, Yu Gao, Zhigang Sun, Yiru Wang, Jijun Wang, Jinghao Chai, Qian Cao, Yuweng Heng, Hao Jiang, Yunda Dong, et~al.
\newblock Diffvla: Vision-language guided diffusion planning for autonomous driving.
\newblock \emph{arXiv preprint arXiv:2505.19381}, 2025.

\bibitem[Kwon et~al.(2023)Kwon, Li, Zhuang, Sheng, Zheng, Yu, Gonzalez, Zhang, and Stoica]{kwon2023vllm}
Woosuk Kwon, Zhuohan Li, Siyuan Zhuang, Ying Sheng, Lianmin Zheng, Cody~Hao Yu, Joseph~E. Gonzalez, Hao Zhang, and Ion Stoica.
\newblock Efficient memory management for large language model serving with pagedattention.
\newblock In \emph{Proceedings of the ACM SIGOPS 29th Symposium on Operating Systems Principles}, 2023.

\bibitem[Li et~al.(2025{\natexlab{a}})Li, Lin, Galanti, Tu, and Yang]{li2025disco}
Gang Li, Ming Lin, Tomer Galanti, Zhengzhong Tu, and Tianbao Yang.
\newblock Dis{CO}: Reinforcing large reasoning models with discriminative constrained optimization.
\newblock In \emph{The Thirty-ninth Annual Conference on Neural Information Processing Systems}, 2025{\natexlab{a}}.

\bibitem[Li et~al.(2025{\natexlab{b}})Li, Xiong, Guo, Li, Yan, Xu, Zhou, Chen, Sun, Wang, et~al.]{li2025recogdrive}
Yongkang Li, Kaixin Xiong, Xiangyu Guo, Fang Li, Sixu Yan, Gangwei Xu, Lijun Zhou, Long Chen, Haiyang Sun, Bing Wang, et~al.
\newblock Recogdrive: A reinforced cognitive framework for end-to-end autonomous driving.
\newblock \emph{arXiv preprint arXiv:2506.08052}, 2025{\natexlab{b}}.

\bibitem[Li et~al.(2024)Li, Li, Wang, Lan, Yu, Ji, Li, Zhu, Kautz, Wu, et~al.]{li2024hydra}
Zhenxin Li, Kailin Li, Shihao Wang, Shiyi Lan, Zhiding Yu, Yishen Ji, Zhiqi Li, Ziyue Zhu, Jan Kautz, Zuxuan Wu, et~al.
\newblock Hydra-mdp: End-to-end multimodal planning with multi-target hydra-distillation.
\newblock \emph{arXiv preprint arXiv:2406.06978}, 2024.

\bibitem[Liao et~al.(2025{\natexlab{a}})Liao, Chen, Yin, Jiang, Wang, Yan, Zhang, Li, Zhang, Zhang, et~al.]{liao2025diffusiondrive}
Bencheng Liao, Shaoyu Chen, Haoran Yin, Bo Jiang, Cheng Wang, Sixu Yan, Xinbang Zhang, Xiangyu Li, Ying Zhang, Qian Zhang, et~al.
\newblock Diffusiondrive: Truncated diffusion model for end-to-end autonomous driving.
\newblock In \emph{Proceedings of the Computer Vision and Pattern Recognition Conference}, pages 12037--12047, 2025{\natexlab{a}}.

\bibitem[Liao et~al.(2025{\natexlab{b}})Liao, Kong, Wang, Wang, Ye, He, Xu, and Li]{liao2025cotdrive}
Haicheng Liao, Hanlin Kong, Bonan Wang, Chengyue Wang, Wang Ye, Zhengbing He, Chengzhong Xu, and Zhenning Li.
\newblock Cot-drive: Efficient motion forecasting for autonomous driving with llms and chain-of-thought prompting.
\newblock \emph{IEEE Transactions on Artificial Intelligence}, pages 1--15, 2025{\natexlab{b}}.

\bibitem[Liu et~al.(2025{\natexlab{a}})Liu, Wang, Liu, Zeng, Yan, Sun, and Liu]{liu2025dapo}
Jiacai Liu, Chaojie Wang, Chris~Yuhao Liu, Liang Zeng, Rui Yan, Yiwen Sun, and Yang Liu.
\newblock {DAPO} : Improving multi-step reasoning abilities of large language models with direct advantage-based policy optimization.
\newblock In \emph{The Thirty-ninth Annual Conference on Neural Information Processing Systems}, 2025{\natexlab{a}}.

\bibitem[Liu et~al.(2025{\natexlab{b}})Liu, Zhong, Zhang, Guo, Zheng, Wang, Zhao, Liu, Su, Gao, Lin, and Huiyong]{liu2025reasonplan}
Xueyi Liu, Zuodong Zhong, Qichao Zhang, Yuxin Guo, Yupeng Zheng, Junli Wang, Dongbin Zhao, Yun-Fu Liu, Zhiguo Su, Yinfeng Gao, Qiao Lin, and Chen Huiyong.
\newblock Reasonplan: Unified scene prediction and decision reasoning for closed-loop autonomous driving.
\newblock In \emph{9th Annual Conference on Robot Learning}, 2025{\natexlab{b}}.

\bibitem[Liu et~al.(2025{\natexlab{c}})Liu, Chen, Li, Qi, Pang, Du, Lee, and Lin]{liu2025understanding}
Zichen Liu, Changyu Chen, Wenjun Li, Penghui Qi, Tianyu Pang, Chao Du, Wee~Sun Lee, and Min Lin.
\newblock Understanding r1-zero-like training: A critical perspective.
\newblock In \emph{COLM}, 2025{\natexlab{c}}.

\bibitem[LLC(2025)]{waymo2025e2e}
Waymo LLC.
\newblock Vision-based end-to-end driving - 2025: Waymo open dataset.
\newblock \url{https://waymo.com/open/challenges/2025/e2e-driving/}, 2025.

\bibitem[Luo et~al.(2025)Luo, Li, Xu, Lai, Yang, Chen, Luo, Xie, Jiang, Liu, et~al.]{luo2025adathinkdrive}
Yuechen Luo, Fang Li, Shaoqing Xu, Zhiyi Lai, Lei Yang, Qimao Chen, Ziang Luo, Zixun Xie, Shengyin Jiang, Jiaxin Liu, et~al.
\newblock Adathinkdrive: Adaptive thinking via reinforcement learning for autonomous driving.
\newblock \emph{arXiv preprint arXiv:2509.13769}, 2025.

\bibitem[Parashar et~al.(2025)Parashar, Gui, Li, Ling, Vemuri, Olson, Li, Zhang, Caverlee, Kalathil, et~al.]{parashar2025curriculum}
Shubham Parashar, Shurui Gui, Xiner Li, Hongyi Ling, Sushil Vemuri, Blake Olson, Eric Li, Yu Zhang, James Caverlee, Dileep Kalathil, et~al.
\newblock Curriculum reinforcement learning from easy to hard tasks improves llm reasoning.
\newblock \emph{arXiv preprint arXiv:2506.06632}, 2025.

\bibitem[Philion et~al.()Philion, Peng, and Fidler]{philiontrajeglish}
Jonah Philion, Xue~Bin Peng, and Sanja Fidler.
\newblock Trajeglish: Traffic modeling as next-token prediction.
\newblock In \emph{The Twelfth International Conference on Learning Representations}.

\bibitem[Qian et~al.(2025)Qian, Jiang, Zhong, Luo, Huang, Zhu, Jiang, Yang, Fu, Miao, et~al.]{qian2025agentthnk}
Kangan Qian, Sicong Jiang, Yang Zhong, Ziang Luo, Zilin Huang, Tianze Zhu, Kun Jiang, Mengmeng Yang, Zheng Fu, Jinyu Miao, et~al.
\newblock Agentthink: A unified framework for tool-augmented chain-of-thought reasoning in vision-language models for autonomous driving.
\newblock \emph{arXiv preprint arXiv:2505.15298}, 2025.

\bibitem[Renz et~al.(2025)Renz, Chen, Arani, and Sinavski]{renz2025simlingo}
Katrin Renz, Long Chen, Elahe Arani, and Oleg Sinavski.
\newblock Simlingo: Vision-only closed-loop autonomous driving with language-action alignment.
\newblock In \emph{Proceedings of the IEEE/CVF Conference on Computer Vision and Pattern Recognition (CVPR)}, pages 11993--12003, 2025.

\bibitem[Rowe et~al.(2025)Rowe, de~Schaetzen, Girgis, Pal, and Paull]{rowe2025poutine}
Luke Rowe, Rodrigue de Schaetzen, Roger Girgis, Christopher Pal, and Liam Paull.
\newblock Poutine: Vision-language-trajectory pre-training and reinforcement learning post-training enable robust end-to-end autonomous driving.
\newblock \emph{arXiv preprint arXiv:2506.11234}, 2025.

\bibitem[Sheng et~al.(2024)Sheng, Zhang, Ye, Wu, Zhang, Zhang, Peng, Lin, and Wu]{sheng2024verl}
Guangming Sheng, Chi Zhang, Zilingfeng Ye, Xibin Wu, Wang Zhang, Ru Zhang, Yanghua Peng, Haibin Lin, and Chuan Wu.
\newblock Hybridflow: A flexible and efficient rlhf framework.
\newblock \emph{arXiv preprint arXiv: 2409.19256}, 2024.

\bibitem[Song et~al.(2025)Song, Huai, Jiang, Kong, and Luo]{song2025more}
Xurui Song, Shuo Huai, JingJing Jiang, Jiayi Kong, and Jun Luo.
\newblock More than meets the eye? uncovering the reasoning-planning disconnect in training vision-language driving models.
\newblock \emph{arXiv preprint arXiv:2510.04532}, 2025.

\bibitem[Tang et~al.(2026)Tang, Kan, Shan, and Chen]{tang2026planr1}
Xiaolong Tang, Meina Kan, Shiguang Shan, and Xilin Chen.
\newblock Plan-{R1}: Safe and feasible trajectory planning as language modeling.
\newblock In \emph{International Conference on Learning Representations (ICLR)}, 2026.

\bibitem[Wang et~al.(2025{\natexlab{a}})Wang, Song, He, Chen, Pan, Deng, and Gu]{wang2025hmvlm}
Daming Wang, Yuhao Song, Zijian He, Kangliang Chen, Xing Pan, Lu Deng, and Weihao Gu.
\newblock Hmvlm: Multistage reasoning-enhanced vision-language model for long-tailed driving scenarios.
\newblock \emph{arXiv preprint arXiv:2506.05883}, 2025{\natexlab{a}}.

\bibitem[Wang et~al.(2025{\natexlab{b}})Wang, Luo, Bai, Cao, Che, Chen, Chen, Diamond, Ding, Ding, et~al.]{wang2025alpamayo}
Yan Wang, Wenjie Luo, Junjie Bai, Yulong Cao, Tong Che, Ke Chen, Yuxiao Chen, Jenna Diamond, Yifan Ding, Wenhao Ding, et~al.
\newblock Alpamayo-r1: Bridging reasoning and action prediction for generalizable autonomous driving in the long tail.
\newblock \emph{arXiv preprint arXiv:2511.00088}, 2025{\natexlab{b}}.

\bibitem[Wen et~al.(2025)Wen, Zhu, Zhu, Tang, Li, Zhou, Liu, Shen, Peng, and Feng]{wen2025diffusionvla}
Junjie Wen, Yichen Zhu, Minjie Zhu, Zhibin Tang, Jinming Li, Zhongyi Zhou, Xiaoyu Liu, Chaomin Shen, Yaxin Peng, and Feifei Feng.
\newblock Diffusion{VLA}: Scaling robot foundation models via unified diffusion and autoregression.
\newblock In \emph{Forty-second International Conference on Machine Learning}, 2025.

\bibitem[Xie et~al.(2025)Xie, Xu, He, Hwang, Luo, Ji, Lin, Chen, Lu, Leng, Anguelov, and Tan]{xie2025s4driver}
Yichen Xie, Runsheng Xu, Tong He, Jyh-Jing Hwang, Katie~Z Luo, Jingwei Ji, Hubert Lin, Letian Chen, Yiren Lu, Zhaoqi Leng, Dragomir Anguelov, and Mingxing Tan.
\newblock S4-driver: Scalable self-supervised driving multimodal large language model with spatio-temporal visual representation.
\newblock In \emph{IEEE/CVF Conference on Computer Vision and Pattern Recognition (CVPR)}, 2025.

\bibitem[Yang et~al.(2025)Yang, Yu, Wu, Yan, Li, Cheng, Zou, Fang, Cheng, Qiu, et~al.]{yang2025egovla}
Ruihan Yang, Qinxi Yu, Yecheng Wu, Rui Yan, Borui Li, An-Chieh Cheng, Xueyan Zou, Yunhao Fang, Xuxin Cheng, Ri-Zhao Qiu, et~al.
\newblock Egovla: Learning vision-language-action models from egocentric human videos.
\newblock \emph{arXiv preprint arXiv:2507.12440}, 2025.

\bibitem[Yuan et~al.(2025)Yuan, Tang, Luo, Chen, Qian, Sun, Chu, Cai, Zhang, and Li]{yuan2025autodrive}
Zhenlong Yuan, Jing Tang, Jinguo Luo, Rui Chen, Chengxuan Qian, Lei Sun, Xiangxiang Chu, Yujun Cai, Dapeng Zhang, and Shuo Li.
\newblock Autodrive-r$^2$: Incentivizing reasoning and self-reflection capacity for vla model in autonomous driving.
\newblock \emph{arXiv preprint arXiv:2509.01944}, 2025.

\bibitem[Yue et~al.(2025)Yue, Chen, Lu, Zhao, Wang, Yue, Song, and Huang]{yue2025does}
Yang Yue, Zhiqi Chen, Rui Lu, Andrew Zhao, Zhaokai Wang, Yang Yue, Shiji Song, and Gao Huang.
\newblock Does reinforcement learning really incentivize reasoning capacity in {LLM}s beyond the base model?
\newblock In \emph{The Thirty-ninth Annual Conference on Neural Information Processing Systems}, 2025.

\bibitem[Zeng et~al.(2025)Zeng, Chang, Xie, Liu, Bai, Pan, Xu, and Wei]{zeng2025futuresightdrive}
Shuang Zeng, Xinyuan Chang, Mengwei Xie, Xinran Liu, Yifan Bai, Zheng Pan, Mu Xu, and Xing Wei.
\newblock Futuresightdrive: Thinking visually with spatio-temporal cot for autonomous driving.
\newblock In \emph{The Thirty-ninth Annual Conference on Neural Information Processing Systems}, 2025.

\bibitem[Zhang and Zuo(2025)]{zhang2025grpolead}
Jixiao Zhang and Chunsheng Zuo.
\newblock {GRPO}-{LEAD}: A difficulty-aware reinforcement learning approach for concise mathematical reasoning in language models.
\newblock In \emph{Proceedings of the 2025 Conference on Empirical Methods in Natural Language Processing}, pages 5642--5665, Suzhou, China, 2025. Association for Computational Linguistics.

\bibitem[Zheng et~al.(2025)Zheng, Mao, Ye, Li, Zhan, Lang, and Zhao]{zheng2025driveagentr1}
Weicheng Zheng, Xiaofei Mao, Nanfei Ye, Pengxiang Li, Kun Zhan, Xianpeng Lang, and Hang Zhao.
\newblock Driveagent-r1: Advancing vlm-based autonomous driving with active perception and hybrid thinking.
\newblock \emph{arXiv preprint arXiv:2507.20879}, 2025.

\bibitem[Zhou et~al.(2025{\natexlab{a}})Zhou, Ma, Liang, Shen, Cui, and Zhang]{zhou2025daro}
Jingyu Zhou, Lu Ma, Hao Liang, Chengyu Shen, Bin Cui, and Wentao Zhang.
\newblock Daro: Difficulty-aware reweighting policy optimization.
\newblock \emph{arXiv preprint arXiv:2510.09001}, 2025{\natexlab{a}}.

\bibitem[Zhou et~al.(2025{\natexlab{b}})Zhou, Cai, Zhao, Zhang, Huang, Zhou, and Ma]{zhou2025autovla}
Zewei Zhou, Tianhui Cai, Seth~Z. Zhao, Yun Zhang, Zhiyu Huang, Bolei Zhou, and Jiaqi Ma.
\newblock Auto{VLA}: A vision-language-action model for end-to-end autonomous driving with adaptive reasoning and reinforcement fine-tuning.
\newblock In \emph{The Thirty-ninth Annual Conference on Neural Information Processing Systems}, 2025{\natexlab{b}}.

\end{thebibliography}
% }

% WARNING: do not forget to delete the supplementary pages from your submission 
\clearpage
\setcounter{page}{1}
\maketitlesupplementary
% \section{Comparison between GRPO versus Dr.~GRPO}
% As explained in 
% We present the component-wise breakdown of results in \cref{tab:grpo_drgrpo_comparison} in \cref{tab:navsim_detailed_comparison}.  Further, as evident with the training and validation plots in \cref{fig:train_val_curves}, while the training and validation curves increase for both GRPO and Dr.~GRPO, GRPO almost always lags behind Dr.~GRPO. We also visualize the change in the mean PDM scores of the group for GRPO and Dr.~GRPO, with respect to the SFT model (or equivalently step 0),  across different variance groups in \cref{fig:contour_plot}. The variance groups are created as per intra-group tertiles. We find that for:
% \begin{enumerate}
%     \item \textbf{Low variance samples} (\cref{fig:contour_plot}~(a)) GRPO has higher density above $y=x$ line, especially for higher values of initial score in $[0.8,1.0]$  
%     \item  \textbf{Medium and high variance samples} (\cref{fig:contour_plot}~(b,c)) Dr.~GRPO performs significantly better than GRPO, with significantly concentrated cluster density above the $y=x$ line. The gap is wider for high variance samples, which is consistent with our findings that GRPO attenuates the policy updates for high variance samples. 
% \end{enumerate}
\section{Comparison between GRPO and Dr.~GRPO}
\begin{figure*}
    \centering
    \includegraphics[width=1\linewidth]{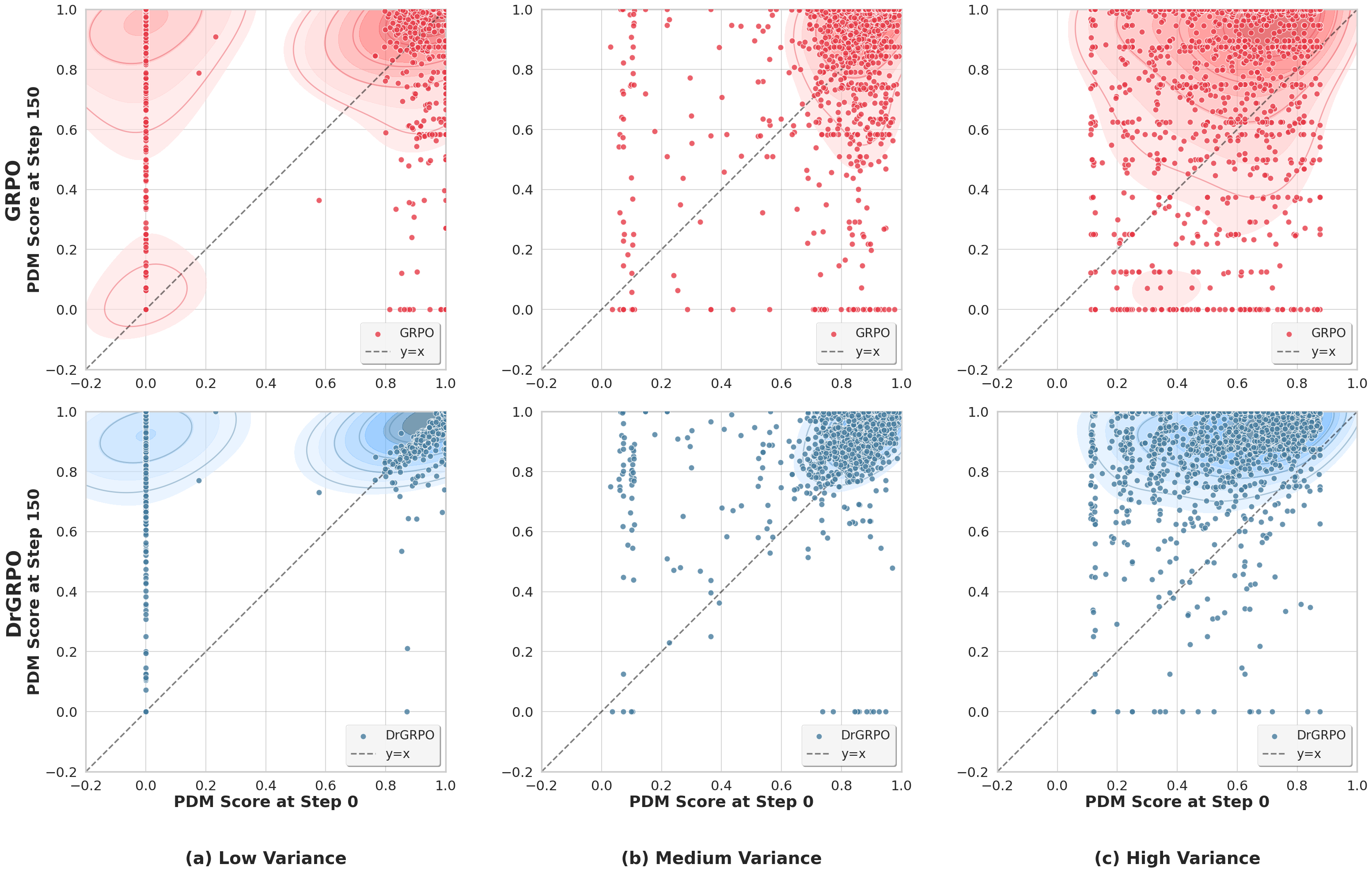}
    \caption{\textbf{Training improvement patterns for GRPO (top, red) and Dr.~GRPO (bottom, blue) across intra-group variance levels.} The $y=x$ line indicates no change in PDM score. GRPO shows strong improvements for low-variance samples with initial scores in $[0.8,1.0]$ (panel (a)), while Dr.~GRPO outperforms GRPO for medium- and high-variance samples (panels (b) and (c)), with denser concentration above $y=x$.}
    \label{fig:contour_plot}
\end{figure*}

\begin{figure}
    \centering
    \includegraphics[width=\linewidth]{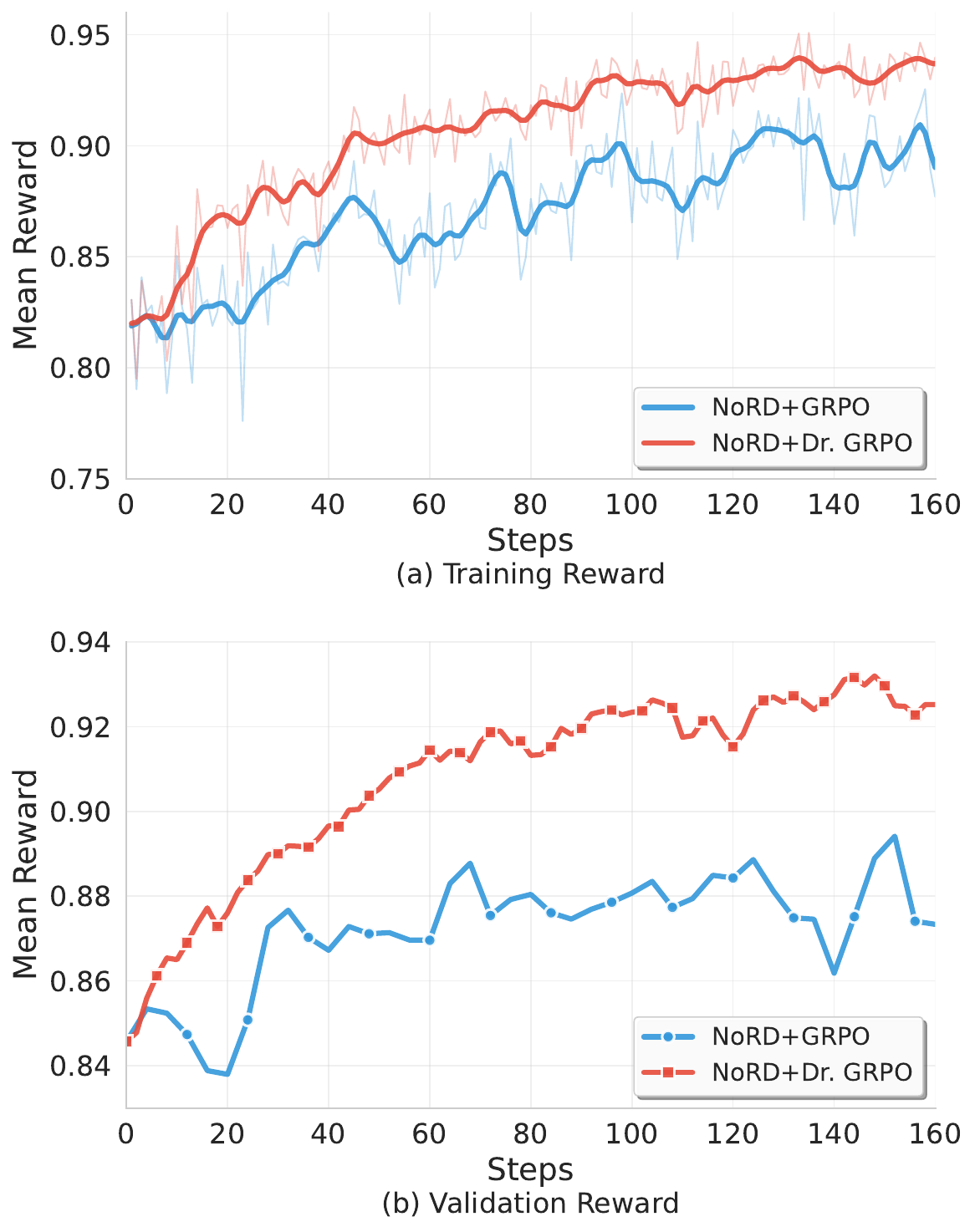}
    \caption{\textbf{Training and validation curves for RL fine-tuning with GRPO and Dr.GRPO.} Dr.GRPO (in red) consistently outperforms GRPO (in blue) on the (a) training and (b) validation sets by a significant margin.}
    \label{fig:train_val_curves}
\end{figure}

We present a component-wise breakdown of  \cref{tab:grpo_drgrpo_comparison} in \cref{tab:navsim_detailed_comparison}. Except for Ego Progress, Dr.~GRPO significantly outperforms GRPO. As shown in the training and validation curves in \cref{fig:train_val_curves}, both GRPO and Dr.~GRPO improve over time; however, GRPO consistently lags behind Dr.~GRPO. To further illustrate this, we visualize the change in mean PDM scores of the group, relative to the SFT model (step 0), across different variance groups in \cref{fig:contour_plot}. The variance groups are defined based on intra-group tertiles. Our analysis reveals that:
\begin{enumerate}
    \item \textbf{Low-variance samples} (\cref{fig:contour_plot}~(a)): GRPO exhibits higher density above the $y=x$ line, particularly for initial scores in $[0.8,1.0]$.
    \item \textbf{Medium- and high-variance samples} (\cref{fig:contour_plot}~(b,c)): Dr.~GRPO outperforms GRPO, with a denser concentration above the $y=x$ line. The performance gap widens for high-variance samples, consistent with our observation that GRPO attenuates policy updates for such samples.
\end{enumerate}

\begin{table*}
\caption{Detailed comparison of RL-fine-tuning of {\modelname{\textsc{-base}}} with GRPO and Dr.~GRPO. Dr.~GRPO based RL fine-tuning is almost always better than GRPO.}
\label{tab:navsim_detailed_comparison}
\centering
\begin{tabular}{@{}lcccccccc@{}}
\toprule
% Method & w/o R & w/o L & C & PDMS$\uparrow$ & Collision$\uparrow$  & DAC$\uparrow$  & Direction$\uparrow$  & Progress$\uparrow$  & TTC$\uparrow$  & Comfort$\uparrow$ \\
\textbf{Method} & \textbf{PDMS}$\uparrow$ & \textbf{Collision}$\uparrow$  & \textbf{DAC}$\uparrow$  & \textbf{Direction}$\uparrow$  & \textbf{Progress}$\uparrow$  & \textbf{TTC}$\uparrow$  & \textbf{Comfort}$\uparrow$ \\
\midrule
{\modelname{\textsc{-base}}} & 76.66 &	96.45 &	86.37	& 94.62	& 71.58	& 90.37	& 99.97\\
{\modelname{\textsc{-base}}{{+GRPO}}} & 77.18 &	91.89 &	90.12	& 91.84	 & \textbf{80.06}	& 80.13	& 99.96 \\
{\modelname{\textsc{-base}}{{+Dr.~GRPO}}} & \textbf{85.62} & \textbf{97.56} & \textbf{94.92} & \textbf{95.94} & 79.30 & \textbf{93.53} & \textbf{100} \\
\bottomrule
\end{tabular}
\end{table*}

\section{Detailed Results}

\subsection{Prompt Example}
\begin{figure}
    \centering
    \includegraphics[width=1\linewidth]{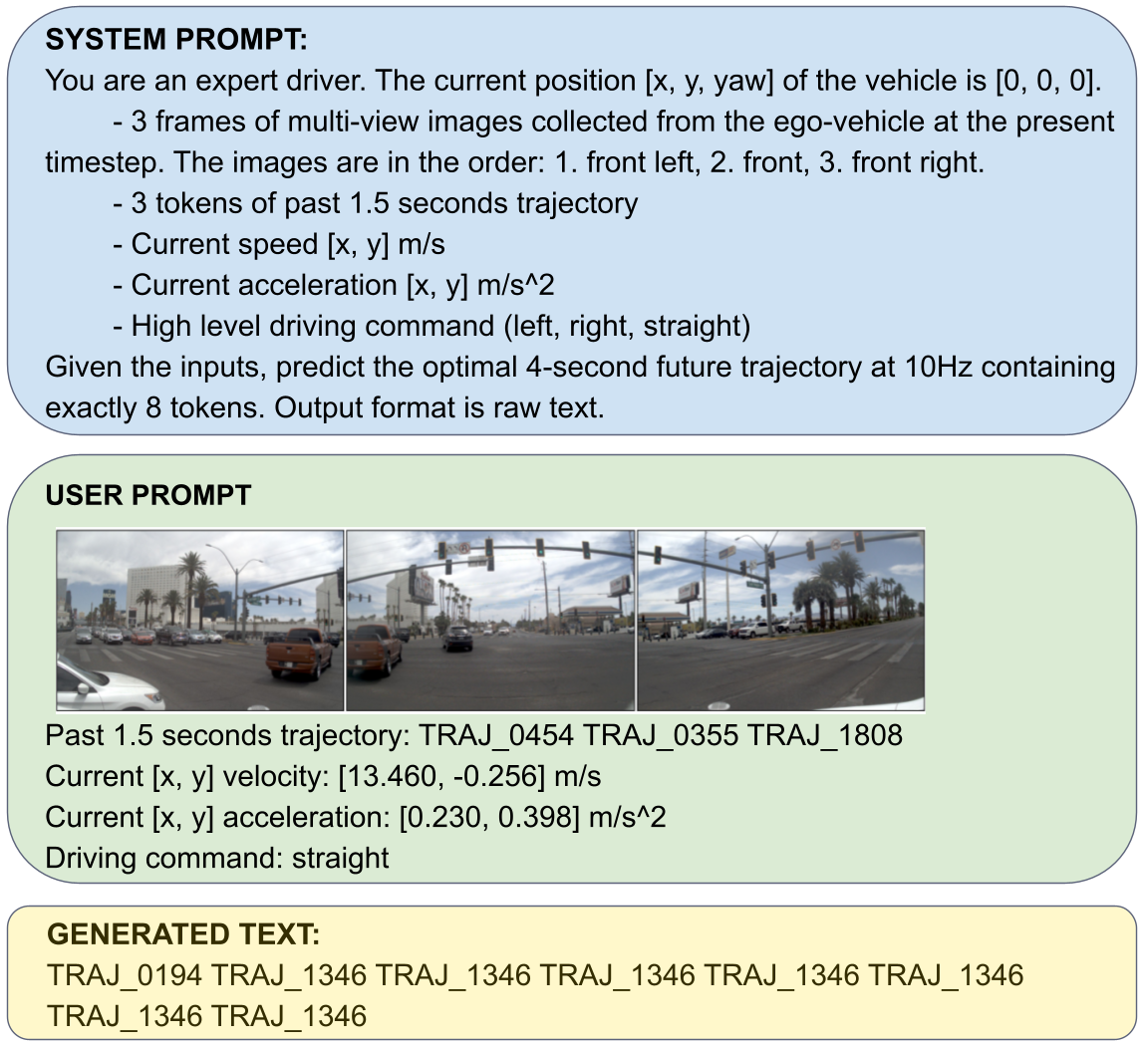}
    \caption{\textbf{Example of \modelname inference.} Given multi-view images, past trajectory, and the current velocity, acceleration, and driving command, \modelname directly predicts the trajectory tokens without explicit reasoning.}
    \label{fig:prompt_example}
\end{figure}

We show an illustrative example in \cref{fig:prompt_example}. \modelname maintains token and inference efficiency by directly predicting the trajectory tokens.

\subsection{Waymo E2E Scores}
We present the detailed results of the performance of \modelname on WaymoE2E test set in \cref{tab:waymo_detailed}. As is evident, \modelname is capable of performing complex multi-lane switching maneuvers, while also performing well in less-represented scenes, such as intersections and construction sites.

\subsection{Effect of Vocabulary Size}
We experimented with a smaller k-disc vocabulary, consisting of 512 trajectory tokens (as compared to 2048 trajectory tokens in \modelname) and found that the performance on NAVSIM degrades (\cref{tab:vocab_size}). This is perhaps because the smaller vocabulary size cannot represent complex maneuvers like sharp turn faithfully.
\begin{table}
\caption{Effect of k-disc vocabulary size on the performance of \modelname on navtest.}
\label{tab:vocab_size}
\centering
\begin{tabular}{@{}ll@{}}
    \toprule
    \textbf{Vocabulary Size} & \textbf{PDMS} $\uparrow$ \\
    \midrule
    512 & 83.07 \\
    2048 & 85.62 \\
    \bottomrule
 \end{tabular}
\end{table}

\begin{table}
\caption{Detailed results on WaymoE2E Test Set.}
\label{tab:waymo_detailed}
\centering
\begin{tabular}{@{}lc@{}}
    \toprule
    \textbf{Metric Name} & \textbf{Value$\uparrow$} \\
    \midrule
    Construction Score & 8.072616 \\
    Intersection Score & 7.9252014 \\
    Pedestrian Score & 7.7775736 \\
    Cyclist Score & 7.8055406 \\
    Multi Lane Maneuver Score & 7.8262477 \\
    Single Lane Maneuver Score & 8.308635 \\
    Cut In Score & 7.734755 \\
    Foreign Object Debris Score & 7.6988134 \\
    Special Vehicle Score & 7.7961473 \\
    Spotlight Score & 6.5309787 \\
    Others Score & 7.322814 \\
    ADE at 3 seconds & 1.250462 \\
    ADE at 5 seconds & 2.8928785 \\
    \midrule
    \textbf{Average Score} & 7.709029 \\
    \bottomrule
\end{tabular}
\end{table}

\section{Reward Functions}
In this section, we elaborate on the reward functions used for RL post-training. The reward consists of length reward, format reward and dataset-specific reward (PDM score for NAVSIM and Normalized RFS for WaymoE2E). The output of the model is a string of action tokens like \texttt{TRAJ\_0242 TRAJ\_150 TRAJ\_172} that are decoded to a list of waypoints of tuples \texttt{[x,y,yaw]} at 10~Hz.
\newline
\newline
\noindent\textbf{Format Reward $(r_f)$:} A binary reward taking values in $\{0,\,0.25\}$. A reward of $0.25$ is assigned if the prediction consists of valid space-separated trajectory tokens of the form \texttt{TRAJ\_i}, where $i$ is a zero-padded 4-digit integer in $[0, 2047]$; otherwise the reward is 0.
\newline
\newline
\noindent\textbf{Length Reward $(r_l)$:} A binary reward taking values in $\{0,\,0.25\}$. The model receives a reward of $0.25$ if the prediction contains the correct number of trajectory tokens (8 for NAVSIM and 10 for WaymoE2E); otherwise, the reward is 0.
\newline
\newline
\noindent\textbf{Dataset Specific Reward $(r_d)$}: 
\begin{enumerate}
    \item \textbf{PDM Score for NAVSIM}: The PDM score (range: $[0,1]$) comprehensively measures the driving quality and safety. Is it given by:
    \[
    \textrm{PDM Score} = \textrm{NC} \times \textrm{DAC} \times \frac{5\cdot\textrm{TTC} + 2 \cdot \textrm{C} + 5 \cdot \textrm{EP}}{12}
    \]
    where, No at-fault Collision (NC), Drivable Area Compliance (DAC), Ego Progress (EP), Comfort (C), and Time-to-Collision (TTC) are all within $[0,1]$. 
    
    \item \textbf{Normalized RFS for WaymoE2E}: The RFS quantifies the alignment of the model's predicted trajectory $\hat{T}$ with a set of three pre-rated human trajectories $T_r$. A score $s_r \in [3,10]$ is assigned to each rater trajectory based on whether $\hat{T}$ falls within a \textit{trust region} defined by dynamic longitudinal $\bar{\tau}_{\text{lng}}$ and lateral $\bar{\tau}_{\text{lat}}$ thresholds (scaled by current velocity). The final score is $\max_r\left(s_r\right)$, averaged over $t \in \{3, 5\}$ seconds, and clipped to $\min(\cdot, 4)$. The Normlized RFS, with range $[0,1]$ is then given by:
    \[
    \textrm{Normalized RFS} = \frac{\max(\max_r (s_r), 4) - 4}{6}
    \]
\end{enumerate}

\noindent The overall reward $r$ for the predicted trajectory is therefore given as: 
\[
r = \frac{r_f + r_l + r_d}{1.5}
\]

\section{Dataset Details}
\subsection{WaymoE2E}
\noindent\textbf{Supervised Finetuning:} We curated the SFT dataset from the official WaymoE2E training set. Frames were first strictly filtered, retaining only those that guaranteed four preceding time steps were available for consistent extraction of the ego-vehicle's historical states. The final subset was then created by uniformly sampling $20\%$ of these valid frame sequences from all contexts. This dataset was then randomly split into training and validation sets using an $85/15$ ratio. The input images were resized to ensure the total number of pixels lies between 784 and 401,408, following the Qwen vision encoder’s constraints.
\newline
\newline
\noindent\textbf{RL Finetuning:} We use the official WaymoE2E validation set, for which preference annotations are provided for a single frame per scenario. Consequently, we extract one sample per scenario and randomly split the resulting set into training and validation sets using an $85/15$ ratio.

\subsection{NAVSIM}
\noindent\textbf{Supervised Finetuning:} We use the official NAVSIM's training set (\texttt{navtrain}) and split it into training and validation sets for SFT using an $80/20$ ratio. The input images were resized to ensure the total number of pixels lies between 784 and 401,408, following the Qwen vision encoder’s constraints.
\newline
\newline
\noindent\textbf{RL Finetuning:} We construct a RLFT dataset from the NAVSIM validation split originally used for supervised fine-tuning. To remove trivial driving behaviors, we filter trajectories using a constant-velocity baseline and discard samples with a final-point displacement error below 0.2 m. For turning maneuvers, we additionally enforce a minimum average heading change of 0.01 rad per timestep to eliminate mild curvature and drift. Straight trajectories are exempt from the heading filter and are filtered solely using the displacement criterion. After filtering, the remaining samples are balanced across three driving intents—straight, left, and right—by uniformly subsampling each class. The resulting dataset contains only non-trivial and dynamically diverse trajectories, providing a more rigorous training signal for reinforcement learning-based trajectory prediction and decision-making models.

\section{Implementation Details}
\subsection{Supervised Finetuning}
We perform supervised fine-tuning of \modelname on the NAVSIM and WaymoE2E datasets using the {Qwen2.5-VL-3B-Instruct} backbone, adapted to predict discretized trajectory tokens from multi-view images, past trajectories, and the ego-vehicle's current kinematic states. For NAVSIM, inputs consist of three camera frames (Front-Left, Front, Front-Right), three past trajectory tokens covering the previous 1.5 seconds, current velocity and acceleration, and a high-level driving command, with the model predicting 8 future trajectory tokens over a 4-second horizon at 10Hz. For WaymoE2E, inputs include six past trajectory tokens spanning 3 seconds, and the model predicts 10 future tokens over a 5-second horizon. In both cases, trajectory tokens are incorporated into the model vocabulary. All components of the model, including the vision encoder, multimodal MLP, and language model, are fine-tuned using mixed-precision training with bf16 and gradient checkpointing to reduce memory footprint. We train the model across 16 A100 GPUs, applying DeepSpeed ZeRO Stage 3 optimization for WaymoE2E and standard distributed training for NAVSIM. We use consistent hyperparameters across datasets, including a learning rate of $5 \times 10^{-5}$, a batch size of 8 per device with 4 gradient accumulation steps, a cosine learning rate scheduler, a warmup ratio of 0.03, and gradient clipping at 1. We evaluate the model every 50 steps on the validation sets and select the best model based on minimum evaluation loss. 

\subsection{RL Finetuning}

We perform RL fine-tuning of \modelname using Dr.~GRPO to optimize task-specific rewards. We generate 8 rollouts per input to estimate group-relative advantages and update the policy accordingly. We use a batch size of 128 trajectories for NAVSIM and 256 trajectories for WaymoE2E, applying asymmetric clipping with a high clip of 0.1 and a low clip of -0.2 to stabilize policy updates. We train across 32 A100 GPUs for WaymoE2E and 30 A100 GPUs for NAVSIM, leveraging mixed-precision and gradient checkpointing for memory efficiency. We periodically evaluate the policy on validation sets and retain the checkpoint achieving the highest reward. 

\section{Dataset Scale Estimation} 
To visualize the performance-efficiency frontier in \cref{fig:pareto_combined}, we estimated the total number of training samples for all evaluated models based on their reported configurations. Across both NAVSIM and WaymoE2E, baseline methods frequently employ complex multi-dataset mixtures or utilize varying fractions of the available data. To standardize these counts, we explicitly aggregated the reported dataset percentages and official splits detailed in the respective papers' training sections. For example, on WaymoE2E, we calculate HMVLM and DiffusionLTF at approx. 500k and 730k samples based on the train and val splits in the Waymo Open Dataset for end-to-end driving and perception. Similarly, for Poutine and AutoVLA, we aggregate their reported multi-dataset percentages to approx. 700k and 210k samples, respectively. These standardizations ensure a fair relative comparison of data efficiency on the x-axis.

\section{Failure Cases}
\begin{figure}
    \centering
    \includegraphics[width=\linewidth]{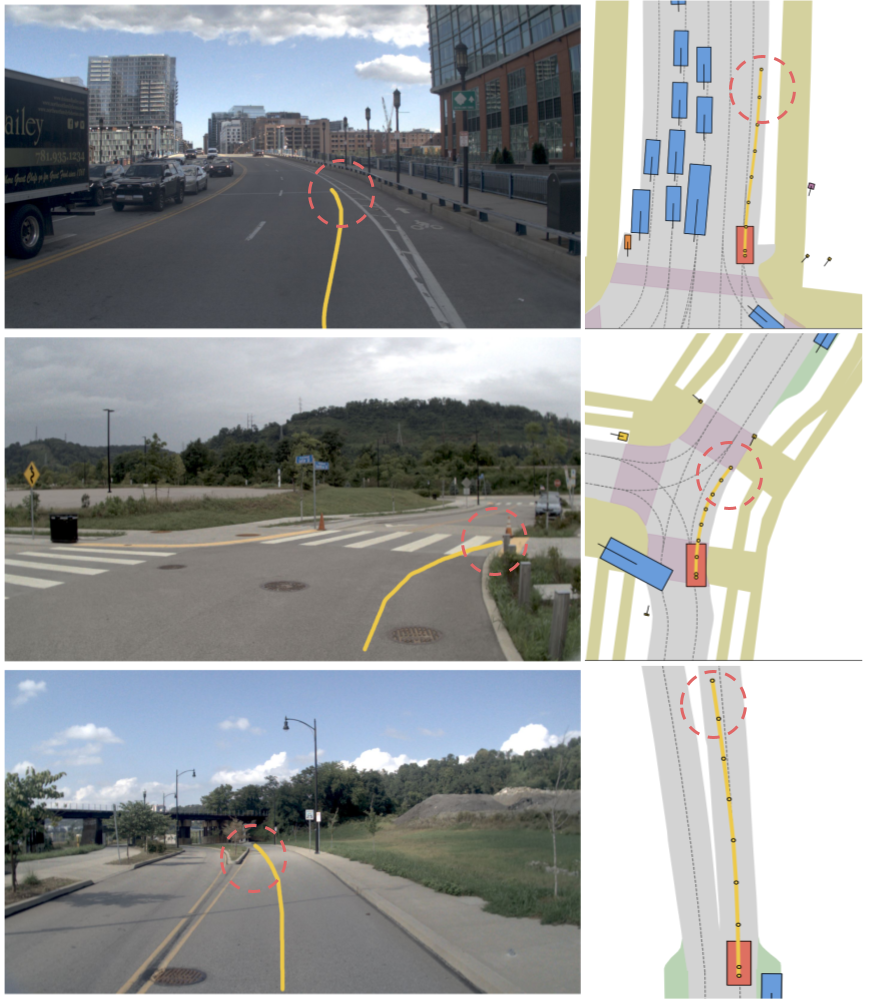}
    \caption{\textbf{Failure cases of \modelname.} The predicted trajectory is shown in red and the violations marked in red circle.}
    \label{fig:failures}
\end{figure}

While \modelname achieves strong performance, it remains susceptible to failure in certain scenarios. We present representative examples in \cref{fig:failures}. These cases can be attributed, in part, to the fact that Dr.~GRPO remains susceptible to difficulty bias, which still affects the policy optimization dynamics. We therefore believe that targeted interventions to better account for task difficulty could further push the performance frontier.

% \section{Rationale}
% \label{sec:rationale}
% % 
% Having the supplementary compiled together with the main paper means that:
% % 
% \begin{itemize}
% \item The supplementary can back-reference sections of the main paper, for example, we can refer to \cref{sec:intro};
% \item The main paper can forward reference sub-sections within the supplementary explicitly (e.g. referring to a particular experiment); 
% \item When submitted to arXiv, the supplementary will already included at the end of the paper.
% \end{itemize}
% % 
% To split the supplementary pages from the main paper, you can use \href{https://support.apple.com/en-ca/guide/preview/prvw11793/mac#:~:text=Delete%20a%20page%20from%20a,or%20choose%20Edit%20%3E%20Delete).}{Preview (on macOS)}, \href{https://www.adobe.com/acrobat/how-to/delete-pages-from-pdf.html#:~:text=Choose%20%E2%80%9CTools%E2%80%9D%20%3E%20%E2%80%9COrganize,or%20pages%20from%20the%20file.}{Adobe Acrobat} (on all OSs), as well as \href{https://superuser.com/questions/517986/is-it-possible-to-delete-some-pages-of-a-pdf-document}{command line tools}.

\end{document}